\documentclass[pdflatex,sn-nature]{sn-jnl}

\usepackage{graphicx}%
\usepackage{amsmath,amssymb,amsfonts}%
\usepackage{xcolor}%
\usepackage{booktabs}%
\usepackage{siunitx}%
\usepackage[normalem]{ulem}
\unnumbered

\begin{document}

\title[Accurate and Auditable ECG Classification]{Accurate and Auditable ECG Classification via Visibility Graph Neural Networks}

\author*[1]{\fnm{Artur} \sur{Xavier}}\email{arturxavier@dcc.ufmg.br}
\author[1]{\fnm{Guilherme Henrique Gomes} \sur{Evangelista}}
\author[2]{\fnm{Aitazaz} \sur{Khan}}
\author[3]{\fnm{Antonio Luiz Pinho} \sur{Ribeiro}}
\author[2]{\fnm{Mohammed J.} \sur{Zaki}}
\author*[4]{\fnm{Srinivasan} \sur{Parthasarathy}}\email{srini@cse.ohio-state.edu}
\author*[1]{\fnm{Wagner} \sur{Meira~Jr.}}\email{meira@dcc.ufmg.br}

\affil[1]{\orgdiv{Department of Computer Science}, \orgname{Universidade Federal de Minas Gerais}, \orgaddress{\city{Belo Horizonte}, \state{Minas Gerais}, \country{Brazil}}}
\affil[2]{\orgdiv{Department of Computer Science}, \orgname{Rensselaer Polytechnic Institute}, \orgaddress{\city{Troy}, \state{New York}, \country{USA}}}
\affil[3]{\orgdiv{Department of Internal Medicine, School of Medicine}, \orgname{Universidade Federal de Minas Gerais}, \orgaddress{\city{Belo Horizonte}, \state{Minas Gerais}, \country{Brazil}}}
\affil[4]{\orgdiv{Department of Computer Science and Engineering}, \orgname{The Ohio State University}, \orgaddress{\city{Columbus}, \state{Ohio}, \country{USA}}}

\newcommand{\ak}[1]{\textcolor{red}{\textbf{[AK: #1]}}}
\newcommand{\add}[1]{\textcolor{blue}{#1}}
\newcommand{\del}[1]{\textcolor{red}{\sout{#1}}}
\abstract{Physicians read electrocardiograms (ECGs) from waveform shape and timing, yet accurate deep-learning classifiers rarely show which waveform regions drive their predictions. We introduce LYNX-ECG, a graph neural network that represents each heartbeat as a sparse visibility graph of timestamped turning points. On CODE-Test, LYNX-ECG reaches a macro F1 score of 0.889, outperforming five baselines including the ECGFounder foundation model, while its graphs need about 14\% of the storage of graphs built from every sample. Because each node maps to a specific ECG time, we can delete clinically defined waveform regions and alter beat timing to audit predictions. For conduction disorders, the clinically expected region was the strongest evidence in more than 90\% of correct predictions. Bradycardia and tachycardia calls follow heart rate with conduction intervals held fixed, and atrial fibrillation calls vanish when beat-to-beat variation is removed. Sparse visibility graphs thus pair accurate ECG classification with clinically grounded, testable explanations.
}


\maketitle

The 12-lead electrocardiogram (ECG) is a gold-standard, non-invasive test used to detect cardiac rhythm and conduction disorders~\cite{ribeiro2020,wagner2020ptbxl}. Each lead captures the heart's electrical activity over time from a unique perspective. Physicians interpret these waveforms by analyzing their shape, duration, and sequence across beats and leads.  A machine learning model designed to assist in this analysis should be explainable, providing evidence linked to identifiable regions that physicians want to investigate.

Recent machine learning models have demonstrated strong performance in classification tasks by analyzing raw ECG measurements over time~\cite{ribeiro2020,strodthoff2021ptbxlbench,li2024ecgfounder,wang2025melp,zhou2025htuning}. ECGFounder, pretrained on more than \num{10} million ECGs, itself identifies black-box opacity as an open limitation~\cite{li2024ecgfounder}. These models represent clinical structures, such as beat boundaries and waveform landmarks, only implicitly, through raw sample positions and amplitudes. As a result, standard attribution methods often highlight isolated points instead of meaningful waveform regions. Additionally, it remains challenging to establish the model's true dependence on these highlighted areas. Attribution methods, like Integrated Gradients (IG)~\cite{sundararajan2017ig}, are highly sensitive to the choice of reference input~\cite{sturmfels2020baselines,ancona2018towards}. Explanation methods for graph neural networks (GNNs) can disagree with one 
another~\cite{yuan2022survey,ying2019gnnexplainer,luo2020pgexplainer,yuan2021subgraphx}, attention can fail faithfulness tests~\cite{azzolin2025formal}, and redundant predictors can yield seed-dependent explanations~\cite{tai2025redundancy}, so explanations need localization, perturbation, and replication checks~\cite{yang2025tif,zhang2025confexplainer}.

Graph-based representations of ECGs provide a potential solution. However, such an approach (e.g., based on visibility graphs~\cite{lacasa2008,oliveira2024vgecg}) must represent waveform regions while keeping graph construction manageable. Such a graph should preserve the predictive signal and the timestamps needed for regional interventions without assigning a node to every sample. Existing sparse and relational ECG methods choose other units: learned subsequences~\cite{liu2025softshape}, relations between channels~\cite{zhang2026medmamba}, three hand-designed P/QRS/T segment nodes per beat~\cite{lee2025ecggraphnet}, or lead-level graphs for myocardial-infarction localization~\cite{maurer2026xgnn4mi}. Prior ECG visibility graphs operate on raw samples~\cite{oliveira2024vgecg}. Their naive adaptation, which we call \textit{all-points}, makes every sample a node and connects pairs that are geometrically visible, and it exemplifies the scale constraint: on the full CODE training dataset (CODE100), it exceeds one trillion edges and requires more than one terabyte of storage.

\begin{figure*}[t]
  \centering
  \includegraphics[width=\textwidth]{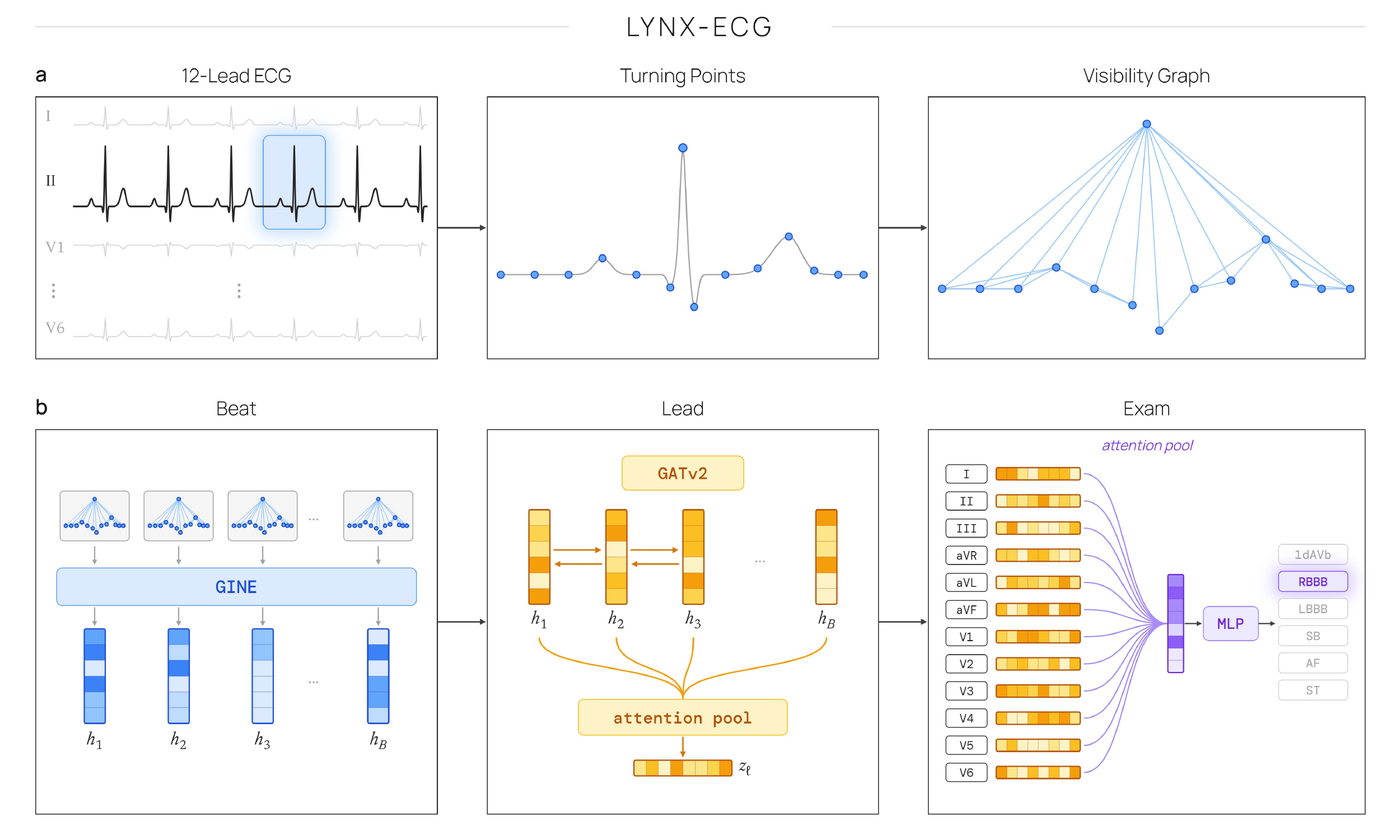}
  \caption{\textbf{The LYNX-ECG pipeline.} (a)~LYNX-ECG keeps the timestamped turning points of each ECG beat and connects them with the natural visibility criterion. (b)~A hierarchical GNN encodes each beat graph with an edge-aware Graph Isomorphism Network (GINE), combines the beats of each lead with a graph attention network (GATv2) and attention pooling, and pools the 12 lead embeddings into one exam embedding. A multilayer perceptron (MLP) head produces the six logits.}
  \label{fig:overview}
\end{figure*}

We explore \textit{turning-point visibility graphs}, an alternative that locally sparsifies the naive graph representation~\cite{satuluri2011local} by selecting samples before graph construction. This selection uses local extrema to capture waveform shape and clinical landmarks, such as R peaks and points
associated with ventricular activation, to help define individual beats. Unlike P/QRS/T segment nodes~\cite{lee2025ecggraphnet}, the retained nodes carry no clinical wave labels. For each beat, the selected samples form the graph nodes, and the natural visibility criterion determines their connections.  We encode each beat graph with a GNN, then combine the resulting representations first within each lead and then across leads to make exam-level predictions. We call the resulting model LYNX-ECG (Lead-level visibilitY Networks for eXplainable ECG). Because each node retains its original timestamp, we can map attribution and deletion results back to predefined ECG landmarks and regions, such as the R peak and other waveform segments. Fig.~\ref{fig:overview} summarizes these steps.

Our contributions are as follows. First, LYNX-ECG adapts visibility graphs to ECG classification and is competitive with ECG foundation models such as ECGFounder. Second, turning-point selection reduces the node count by more than 86\% while keeping this performance. Third, by combining IG with structural deletion, timing interventions, and an exact lead decomposition, we audit which regions, timing, and leads drive each diagnosis, an audit that raw-signal models can approximate only by masking samples. Fourth, we evaluate how node selection, topology, edge attributes, and message direction on two real-world datasets affect classification tasks.

\section{Results}\label{sec:results}

We evaluated turning-point visibility graphs for six-label ECG
classification on held-out CODE-Test and on PTB-XL as an external
dataset~\cite{wagner2020ptbxl}, whose diagnostic labels we mapped to the
same six targets (Methods; Supplementary Table~1).

We report macro F1, the area under the receiver operating characteristic curve (AUROC), and the area under the precision--recall curve (AUPRC). F1, precision, and recall use a fixed threshold of \num{0.5}, and every result covers three training seeds.

\subsection{Competitive classification with smaller graphs}\label{sec:performance}

We compared LYNX-ECG with five baseline configurations
(Table~\ref{tab:performance}; Methods): a deep neural network
(DNN)~\cite{ribeiro2020}, a Vision Transformer (ViT)~\cite{dosovitskiy2021vit},
EfficientNet~\cite{tan2019efficientnet}, and the ECG foundation model
ECGFounder~\cite{li2024ecgfounder}, trained from random initialization
(ECGFounder-Scratch) or fine-tuned from pretrained weights
(ECGFounder-Finetuned).

Our turning-point model was trained from random initialization. On CODE-Test,
it achieved the highest mean macro F1, \(0.889\), and the highest mean AUROC
among the evaluated models. Paired exam-level comparisons also support higher
macro F1 than all five baselines
(Supplementary Table~2).  On PTB-XL, our model reached a macro F1
of \(0.730\), close to both ECGFounder variants. Paired comparisons support
higher F1 than DNN and ViT, but do not establish a difference from EfficientNet
or either ECGFounder variant.  These external
results were not uniform across diagnoses (Extended Data Table~\ref{tab:perclass-f1}). The largest decline at the fixed threshold occurred for first-degree atrioventricular block (1dAVb). Sinus bradycardia (SB) also declined because PTB-XL assigns SB below 60~beats/min, whereas the CODE labels require 50~beats/min or less (Supplementary Note~12 and Supplementary Table~3). Complete class-wise results are reported in Supplementary Notes~3 and~11, Supplementary Tables~4 and~5, and Supplementary Fig.~1.

This performance does not require the dense all-points representation. Turning-point selection removed more than
\num{86}\% of the post-resampling samples on both datasets. On CODE100, this
reduction translated into graphs that required only about \num{14}\% of the storage
used by the all-points graphs (Table~\ref{tab:selector-analysis}).

\begin{table}[t]
  \caption{Six-label classification on CODE-Test and external PTB-XL.
  Macro F1 and AUROC average the six labels; F1 uses the fixed \num{0.5} threshold. PTB-XL labels come from exact SCP-ECG statement codes (strict-SCP; Methods). Values are mean $\pm$ sample SD
  across three training runs. Boldface marks the highest mean and underline marks the second-highest (ties within 0.001).
  }
  \label{tab:performance}
  \centering
  \footnotesize
  \setlength{\tabcolsep}{4pt}
  \begin{tabular}{@{}lcccc@{}}
    \toprule
    & \multicolumn{2}{c}{CODE-Test ($n=827$)}
    & \multicolumn{2}{c}{PTB-XL strict-SCP ($n=\num{21799}$)} \\
    \cmidrule(lr){2-3}\cmidrule(l){4-5}
    Model & Macro F1 $\uparrow$ & AUROC $\uparrow$
          & Macro F1 $\uparrow$ & AUROC $\uparrow$ \\
    \midrule
    LYNX-ECG
      & $\mathbf{0.889\pm0.004}$ & $\mathbf{0.9984\pm0.0001}$
      & \underline{$0.730\pm0.001$} & \underline{$0.9815\pm0.0005$} \\
    DNN~\cite{ribeiro2020}
      & $0.849\pm0.023$ & $0.9976\pm0.0002$
      & $0.718\pm0.003$ & $0.9809\pm0.0008$ \\
    ViT~\cite{dosovitskiy2021vit}
      & $0.864\pm0.011$ & $0.9971\pm0.0006$
      & $0.715\pm0.010$ & $0.9791\pm0.0018$ \\
    EfficientNet~\cite{tan2019efficientnet}
      & \underline{$0.872\pm0.009$} & \underline{$0.9979\pm0.0003$}
      & $0.728\pm0.003$ & $0.9790\pm0.0013$ \\
    ECGFounder-Scratch~\cite{li2024ecgfounder}
      & $0.863\pm0.004$ & \underline{$0.9979\pm0.0003$}
      & \underline{$0.730\pm0.013$} & $\mathbf{0.9818\pm0.0007}$ \\
    ECGFounder-Finetuned~\cite{li2024ecgfounder}
      & $0.858\pm0.035$ & $0.9961\pm0.0016$
      & $\mathbf{0.734\pm0.003}$ & $0.9730\pm0.0010$ \\
    \bottomrule
  \end{tabular}
\end{table}

\subsection{Node selection changes predictive and storage trade-offs}\label{sec:selector-analysis}

This storage reduction depends on which samples are retained. To examine
how sample selection affects the balance between classification and storage, we compared turning points with two alternatives under a matched
protocol (Table~\ref{tab:selector-analysis}; Methods). Uniform sampling
retains \(k=30\) nodes per beat without explicitly selecting local extrema,
whereas all-points graphs keep every sample.

\begin{table}[t]
  \caption{Node selection, predictive performance, and graph size.
  Predictive values are CODE-Test means $\pm$ sample SD across three
  seeds. Graph counts and Hierarchical Data Format 5 (HDF5) storage cover
  the complete CODE100 dataset; B denotes billions, and all-points nodes
  are the resampled signal samples. LYNX-ECG uses turning points. All representations use
  the same model and effective batch size. Their node counts are not identical.}
  \label{tab:selector-analysis}
  \centering
  \footnotesize
  \setlength{\tabcolsep}{3pt}
  \begin{tabular}{@{}lrrrrccc@{}}
    \toprule
    Representation & \shortstack{Nodes\\/beat} & \shortstack{Nodes\\(B)} & \shortstack{Edges\\(B)} & \shortstack{HDF5\\(GB)}
      & Macro F1 & AUROC & Recall \\
    \midrule
    Turning points
      & 34.46 & 9.83 & 29.27 & 164.6
      & $\mathbf{0.889\pm0.004}$ & $0.9984\pm0.0001$
      & $\mathbf{0.914\pm0.016}$ \\
    Uniform \(k=30\)
      & 30.00 & 8.56 & 28.47 & 121.7
      & $0.878\pm0.009$ & $\mathbf{0.9987\pm0.0002}$
      & $0.887\pm0.004$ \\
    \midrule
    All points & 257.97 & 73.58 & 1180.74 & 1175.6
      & $0.870\pm0.011$ & $0.9985\pm0.0002$
      & $0.884\pm0.025$ \\
    \bottomrule
  \end{tabular}
\end{table}

Compared with uniform sampling, turning points achieved higher mean recall
and higher macro F1 in every CODE-Test seed.  Uniform sampling instead achieved higher precision, AUROC, and AUPRC. For 1dAVb, this contrast was particularly clear: uniform sampling
achieved higher AUPRC but missed more positive exams at the chosen threshold (class-wise results in Supplementary Note~4 and Supplementary Table~6).

We next asked whether keeping every sample helps. On
CODE-Test, turning-point graphs achieved higher macro F1 and recall than
all-points graphs, while on PTB-XL the all-points representation yielded
slightly higher F1 (Supplementary Note~5 and Supplementary Table~7).  Beyond requiring less
storage, turning-point graphs were faster to build and used less memory in a construction benchmark (Supplementary Note~7 and Supplementary Table~8).

\subsection{Graph topology and edge attributes interact}\label{sec:topology-factorial}

Having examined which samples should be retained, we next asked how those
samples should be connected.  Using the same
turning-point nodes, we compared natural visibility with a forward temporal
chain, which connects each node only to the next one in time. For each
topology, we trained one model with the time and amplitude differences
between connected nodes, \((\Delta t,\Delta a)\), and another with these edge
attributes set to zero (Methods). This \(2\times2\) design separates connectivity, edge attributes, and their
interaction.

With the derived time and amplitude attributes, visibility achieved higher
macro F1 than the temporal chain in every CODE-Test seed. Removing these
attributes changed the pattern: the temporal chain then achieved higher F1
in every seed (Extended Data Table~\ref{tab:topology-factorial}). As a result, the effect of visibility
cannot be attributed to connectivity alone. Instead, the results
indicate an interaction between the visibility topology and the information
carried by its edges.

\subsection{Effect of message direction on classification}\label{sec:directionality}

We next examined whether allowing edge information to flow in both temporal directions
improves classification, by adding the reverse of every forward edge and
retraining the model (Methods).

This additional information flow did not yield a clear classification
benefit. Paired comparisons did not establish a difference in macro F1 or
AUPRC on either dataset (Supplementary Note~6 and Supplementary Table~9). AUROC favored
forward-only propagation on PTB-XL, with no supported difference on
CODE-Test.

\paragraph*{Takeaway} Turning-point sparsification trades storage for higher CODE-Test recall than uniform sampling, the benefit of visibility depends on its time and amplitude edge attributes, and reverse messaging gives no established gain.

\subsection{Each diagnosis is audited with the evidence that defines it}\label{sec:interp}

\begin{figure*}[t]
  \centering
  \includegraphics[width=\textwidth]{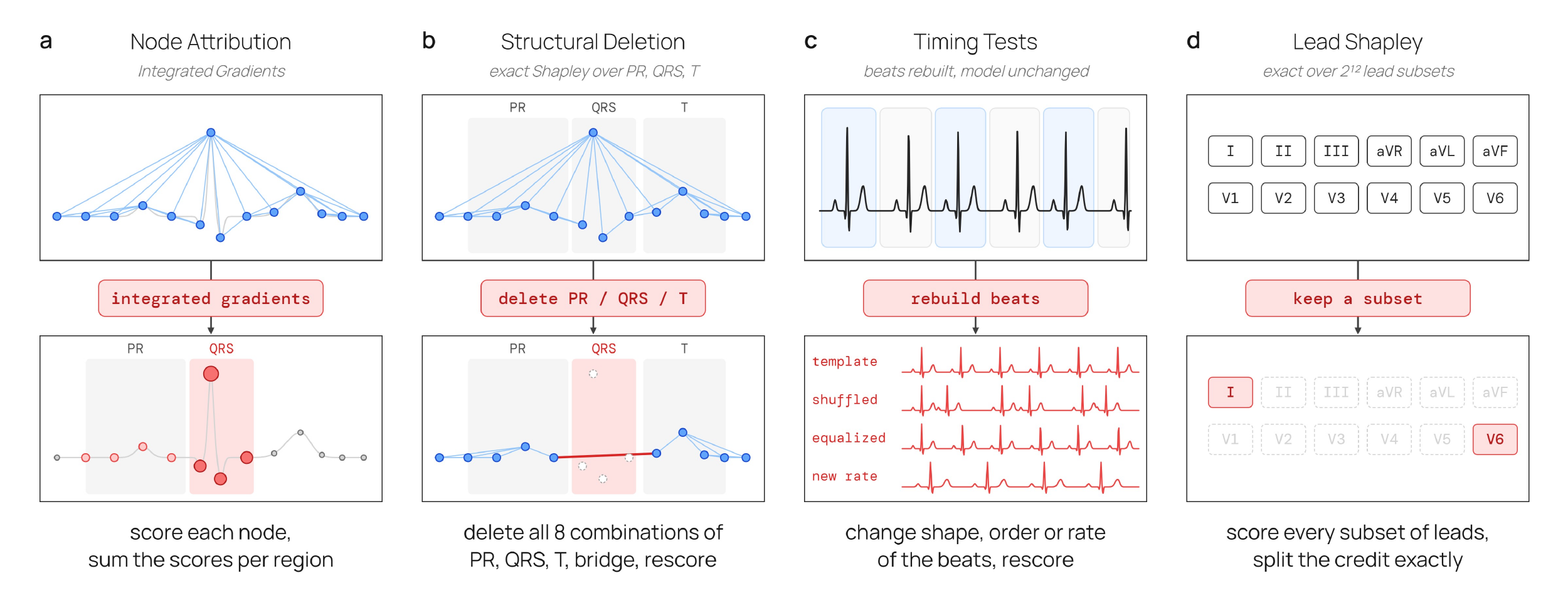}
  \caption{\textbf{Four audits test the evidence behind each prediction.} All audits keep the trained model unchanged. (a)~Node attribution: Integrated Gradients (IG) scores each node, and the scores are summed within each region. (b)~Structural deletion: PR, QRS, and T are removed in every combination, the gap is bridged, and the exam is rescored; exact Shapley values over the three regions give each region's contribution. (c)~Timing tests: the beats are replaced by a template, shuffled, equalized, or rebuilt at a new heart rate, and the exam is rescored. (d)~Lead Shapley: every subset of the 12 leads is scored, and exact Shapley values split the credit among the leads.}
  \label{fig:audits}
\end{figure*}

The six labels differ in the evidence that defines them. First-degree atrioventricular block and right and left bundle-branch block (RBBB, LBBB) are conduction disorders, defined by the duration or shape of regions within each beat. Sinus bradycardia, sinus tachycardia (ST), and atrial fibrillation (AF) are rhythm disorders, defined by rate and regularity across beats. Because nodes keep their timestamps, we delete delineated regions to test conduction labels and change the timing between beats to test rhythm labels. An exact decomposition over the 12 leads then shows which leads carry the evidence for every label. Fig.~\ref{fig:audits} summarizes these tests and the node attribution.

For the conduction labels, we used the PR interval as the expected region for 1dAVb and the QRS complex for both bundle-branch blocks, following the diagnostic criteria for these labels~\cite{ribeiro2020,surawicz2009}. The PR interval spans atrial activation and the conduction delay to the ventricles, which 1dAVb prolongs. A complete bundle-branch block forces one ventricle to activate through a slower path, which widens and distorts the QRS complex. The T wave reflects ventricular recovery. The regional analysis included all 85 CODE-Test exams with at least one of these labels. We evaluated each positive exam--class pair under all three seeds and refer to each exam--class--seed combination as a verdict. The rhythm labels have no expected region; we test them with timing interventions (structural-deletion results in Supplementary Note~15 and Supplementary Table~10).

\subsubsection{Regional attribution depends on its baseline}

\begin{figure*}[t]
  \centering
  \includegraphics[width=\textwidth]{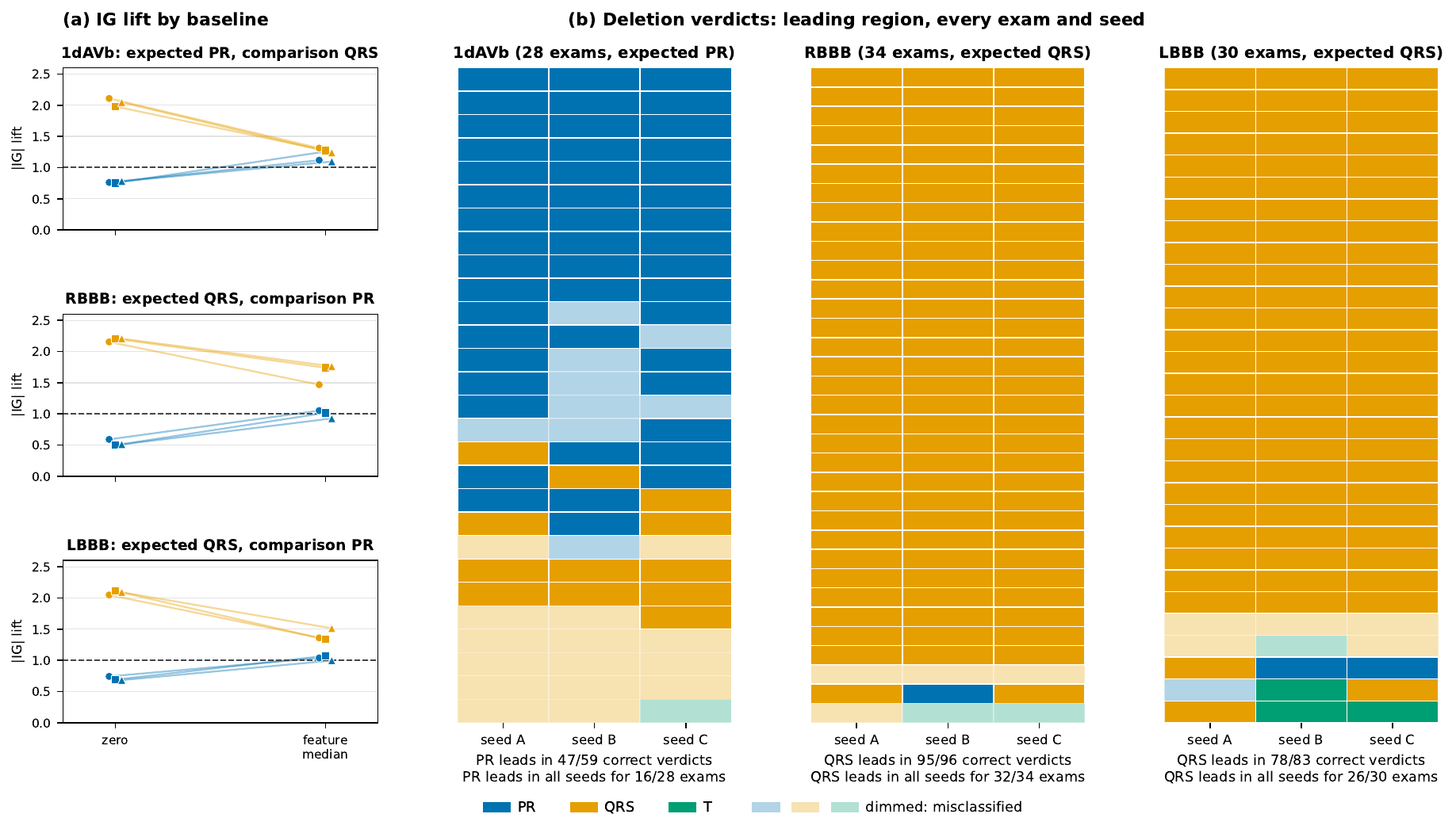}
  \caption{\textbf{Regional attribution and structural-deletion results.}
    (a)~Integrated Gradients (IG) lift of the expected region and of the
    comparison region under the zero and feature-median baselines. PR is the
    expected region for 1dAVb, and QRS is the expected region for RBBB and LBBB. Each
    point is the median over the label-positive CODE-Test exams for one trained seed,
    and a line joins the two baselines of the same seed. A lift of 1 (dashed line)
    means that a region receives attribution in proportion to its number of nodes.
    (b)~Region with the largest Shapley contribution after structural deletion, for
    every label-positive exam (row) and trained seed (column). Dimmed cells are
    misclassified verdicts. The models classify 59 of 84 1dAVb, 96 of 102 RBBB, and
    83 of 90 LBBB verdicts correctly. Across the three labels, the expected region
    leads in 220 of 238 correct verdicts (92.4\%) and in all seeds for 74 of 92
    exam--class pairs, which come from 85 distinct exams.}
  \label{fig:explanation-audit-summary}
\end{figure*}

We first examined whether IG attribution, summarized within fixed windows anchored on each beat's R peak (Methods), was concentrated in the region expected for each diagnosis. IG evaluates
the contribution of each node feature relative to a baseline input. We
compared two baseline choices, an input with features set to zero and an
input using the training-set median of each feature. To compare regions of different
sizes, we used the \emph{lift}, a region's fraction of the total absolute
attribution divided by its fraction of all nodes (Methods). A lift of 1
indicates attribution proportional to the node count. We
compared the expected region with another waveform
region: PR against QRS for 1dAVb, and QRS against PR for RBBB and
LBBB. Extended Data Table~\ref{tab:alignment} and Fig.~\ref{fig:explanation-audit-summary}a report the regional lifts under
both reference inputs.

For RBBB and LBBB, attribution remained concentrated in the expected QRS
region across all three seeds and under both baselines. The pattern was less
stable for 1dAVb. With the zero baseline, PR lift remained below 1, whereas
with the feature-median baseline it rose above 1. In both cases, however,
QRS received greater attribution than PR. This shift suggests that the
regional interpretation for 1dAVb depends on the choice of IG baseline,
consistent with the known sensitivity of attribution methods to their
reference input~\cite{sturmfels2020baselines,ancona2018towards}. Structural deletion tests these regions directly.

\subsubsection{Structural deletion tests regional contributions}

We tested these regional contributions by removing waveform regions
and recomputing the class score. For each exam and target class, we removed the nodes within a selected region and all edges connected to them, and linked the last remaining node before the region to the first remaining node after it, so that each beat graph stayed connected. We did this in every beat where PR, QRS, and T were delineated (Methods) and evaluated all eight
combinations of PR, QRS, and T removal. We used exact Shapley
values~\cite{shapley1953value} to divide the total score drop among the
three regions (Methods). The region with the largest
contribution was considered the \emph{leading region}.

Among correctly classified verdicts, the expected region had the
largest contribution in more than \num{90}\% of cases
(Fig.~\ref{fig:explanation-audit-summary}b). This alignment was strongest for RBBB,
where QRS led in all but one correct verdict, and remained high for LBBB.
For 1dAVb, PR led less often, but the alternative was consistently QRS among correct predictions. Across all 1dAVb verdicts, including misclassified cases, only one placed T first. 

To examine whether this ranking was consistent across training runs, we
counted the exam--class pairs in which the expected region had the largest
contribution under all seeds. This agreement was weaker for 1dAVb than
for the bundle-branch blocks (Fig.~\ref{fig:explanation-audit-summary}b).
Some 1dAVb exams consistently favored QRS, others changed ranking across seeds with closer PR and QRS contributions, and both regions far exceeded T (Supplementary Fig.~2). These findings support
contributions from both PR and QRS, rather than reliance on PR alone. The clinical diagnosis of 1dAVb does not depend on QRS morphology or duration, so the QRS contribution is a property of the model rather than of the diagnostic criterion. Two factors may explain it: the boundary node that ends PR belongs to the QRS segment, and 21\% of 1dAVb exams in the CODE annotations also carry an RBBB or LBBB label, against 4\% of the other exams. Concurrent labels in the test exams do not explain it, because all 12 QRS-led correct verdicts come from exams without a bundle-branch block label (Supplementary Table~11).

To complement the analysis of label-positive exams, we examined the same regions in target-negative exams. In this setting, the expected region had a negative median Shapley contribution for
each class. Removing it therefore raised the class score on average. This pattern matches clinical rule-out reasoning, in which a normal PR interval argues against 1dAVb and a narrow QRS complex argues against a bundle-branch block. The contribution of a region thus depends on the diagnostic context in which it appears.

\subsubsection{Region boundaries affect deletion verdicts}

Structural deletion also depends on how each region is bounded.
We repeated the analysis on the same available beats using two fixed-window definitions anchored on the R peak
(Extended Data Table~\ref{tab:region-ablation}). Compared with per-beat delineation, both fixed-window approaches reduced the number of correct verdicts led by the expected region. The largest differences occurred for RBBB and LBBB. In LBBB, the R-wave peak time in V5--V6 exceeds 60\,ms~\cite{surawicz2009}, so a window that starts 60\,ms before the R peak can omit the QRS onset in the leads that define the diagnosis. For 1dAVb, fixed windows increased the number of T-leading verdicts from one under per-beat delineation to five under either fixed-window protocol.

\subsubsection{Conduction evidence lies within each beat}

\begin{figure*}[t]
  \centering
  \includegraphics[width=\textwidth]{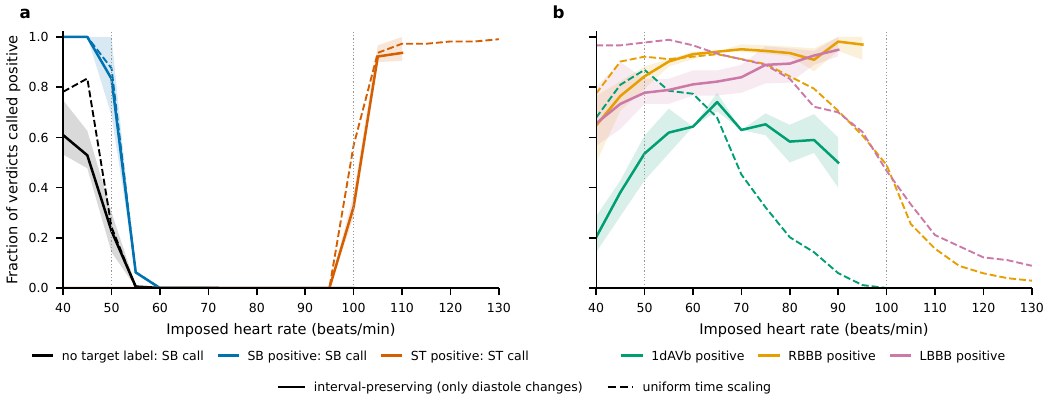}
  \caption{\textbf{Rhythm calls follow heart rate, and conduction calls follow interval duration.}
    Fraction of CODE-Test verdicts called positive as the imposed heart rate changes.
    Dashed lines scale the whole time axis of every beat, which also changes the
    PR, QRS, and QT durations. Solid lines change only the time before P onset and
    after T offset. They are drawn where at least a quarter of the cohort remains
    feasible, because faster rates exhaust the diastole. (a)~Exams without any
    target label called SB, SB-positive exams called SB, and ST-positive exams
    called ST. (b)~Conduction-positive exams called positive. Cohorts are the CODE-Test
    exams without any target label and the label-positive exams (SB \(n=16\),
    ST \(n=37\), 1dAVb \(n=28\), RBBB \(n=34\), LBBB \(n=30\)). Lines are means
    over three seeds, and bands span the seeds.}
  \label{fig:timing}
\end{figure*}

We next asked whether the conduction calls also depend on the beat sequence. For each label-positive verdict, we shuffled the beats within each lead, replaced them with one template beat, or equalized their durations (Methods). The conduction calls barely changed (Supplementary Table~12). With a single template beat, for example, 0.89 of RBBB verdicts and 0.94 of LBBB verdicts stayed positive, against 0.94 and 0.92 intact. The evidence for these labels therefore lies within each beat.

Heart rate is harder to isolate, because stretching or compressing the whole time axis also changes the PR, QRS, and QT durations. Under such a uniformly applied change, the conduction calls fell at faster imposed rates, to 0.49 for RBBB and 0.47 for LBBB at 100~beats/min (Fig.~\ref{fig:timing}b, dashed). We then changed only the pause between the end of one T wave and the next P wave, keeping PR, QRS, and QT durations constant. Under this interval-preserving change, at least 0.92 of RBBB verdicts stayed positive between 60 and 100~beats/min (Fig.~\ref{fig:timing}b, solid; Supplementary Table~13). This shows that the bundle-branch calls follow interval duration, as the QRS criterion requires, and not heart rate. 1dAVb calls declined only moderately between 60 and 80~beats/min, and all conduction calls fell at 40~beats/min.

\subsubsection{Rhythm evidence lies across beats}

The same interventions test the rhythm labels. With the conduction intervals held fixed, half of the exams without any target label were called SB once the imposed rate fell below about 45~beats/min, close to the 50~beats/min criterion of the CODE labels, and ST-positive exams lost the ST call below about 102~beats/min (Fig.~\ref{fig:timing}a; Supplementary Table~13 and Supplementary Fig.~3). No SB-positive verdict kept its call at an imposed 60~beats/min. Rate alone thus creates and removes both sinus-rate calls. The observed errors agree with this reading. On CODE-Test, false SB calls and missed ST exams had median measured rates at the 50 and 100~beats/min thresholds (Supplementary Fig.~4).

Atrial fibrillation behaved differently. Replacing every beat with one template removed every AF call on CODE-Test and 99\% of AF calls on PTB-XL, whereas shuffling the beats left the calls unchanged (Supplementary Table~12). The AF calls therefore depend on the differences between beats rather than on their order, and a record in which every cycle is identical is no longer called AF.

\subsubsection{Lead evidence}

Exact lead Shapley values (Methods; Supplementary Note~14, Supplementary Table~14, and Supplementary Fig.~5) showed that, for the rhythm labels, the evidence was redundant across leads. Several single leads kept the call alone, and removing one lead changed the call in at most 4\% of correct SB, ST, and AF verdicts on CODE-Test and 7\% on PTB-XL. For the conduction labels, the evidence was concentrated in a few leads. Lead~II alone kept 0.97 of correct 1dAVb calls, and V2 alone kept 0.96 of correct LBBB calls. Although the classic criterion for RBBB is the rsR$'$ pattern in V1 and V2, LYNX-ECG relies on I and V6, where the wide terminal S wave is a supporting criterion. Each of these leads alone kept 0.95 of correct RBBB calls, whereas V1 or V2 alone kept none. Similarly, although fibrillatory waves are usually clearest in V1, V1 alone kept almost no AF calls (0.00 on CODE-Test and 0.03 on PTB-XL).
\paragraph*{Takeaway} Explanation audits reveal that
attribution leaves the 1dAVb evidence baseline-dependent, whereas direct
deletion of measured regions yields stable verdicts: the clinical region
leads in more than \num{90}\% of correct verdicts and in all but one correct RBBB verdict, and the
1dAVb evidence stays inside the PR interval and the QRS complex under every seed and both
region protocols. Timing interventions complete the audit for the rhythm labels. SB and ST follow heart rate with the conduction intervals held fixed, the bundle-branch calls follow interval duration rather than rate, and AF calls depend on the differences between beats. All three findings agree with the clinical definitions.

\section{Discussion}\label{sec:discussion}

Turning-point visibility graphs bring predictive performance, efficiency, and regional auditability into the same ECG representation. LYNX-ECG had the highest CODE-Test macro F1 with a fraction of the all-points storage. This reduction does more than make graph construction practical: because the retained nodes preserve their original timestamps, the same representation supports direct interventions on clinically defined waveform regions. On PTB-XL, all-points graphs achieved a slightly higher F1, but turning-point graphs remained competitive while retaining their substantial advantage in graph size.

The value of these regional tests lies in complementing attribution with an explicit intervention. Structural deletion avoids choosing the feature baseline required by IG, while delineation defines region boundaries without consulting the classifier's outputs. Deletion still depends on the region definition. Fixed windows lowered expected-region leadership, mostly for the bundle-branch blocks, possibly because they miss parts of wider QRS complexes (Extended Data Table~\ref{tab:region-ablation}). The intervention therefore needs both an explicit deletion rule and a clearly specified region definition.

The same timestamps also allow time-based interventions, which show that the model reads interval duration for the bundle-branch blocks and rate for the sinus rhythm labels, as the clinical definitions require. The model also does not treat beats in isolation. Replacing every beat with one template removes the atrial fibrillation calls and leaves the conduction calls intact.

These audits also show where the model's evidence differs from the clinical criteria, as for RBBB, which the model reads in leads I and V6 rather than V1 and V2. The model has no cross-lead edges, and none of the six labels is a territorial diagnosis, so these audits cannot test lead-territory patterns such as inferior infarction in leads II, III, and aVF.

External evaluation shows which parts of the representation transfer. Node times are absolute, so the rate rules carry over to PTB-XL: SB recall was 0.97 below 50~beats/min, and ST recall was 0.92 at 100--105~beats/min (Supplementary Table~3). Ranking also transferred, with per-class AUROC between 0.94 and 0.997 on PTB-XL. The F1 losses came mainly from the label protocols and the fixed operating point. For example, 1dAVb kept an AUROC of 0.94 while its F1 fell to 0.44. We did not recalibrate the threshold or adapt the model to PTB-XL, so these results measure the transfer of a fixed CODE model. Other devices, populations, and labeling standards remain untested. Transfer-calibrated prediction-powered inference~\cite{vadlamani2026} could address this gap.

Cardiologists interpret an ECG through a standard nomenclature of waves, intervals, rates, and leads, and an explanation helps them when it uses the same terms, as these audits do: the PR interval and the QRS complex for conduction disorders, heart rate and beat-to-beat variation for rhythm disorders, and the individual leads for all six labels. Together, these results show that turning-point visibility graphs combine competitive ECG classification, a compact representation, and explanation audits that test which waveform regions, timing, and leads drive a prediction.

\section{Methods}\label{sec:methods}

We describe the construction from individual beats to exam-level predictions, followed by the training protocol and the regional explanation audits. Fig.~\ref{fig:overview} and Extended Data Fig.~\ref{fig:construction} summarize the complete pipeline and the beat-level graph construction, and Fig.~\ref{fig:audits} summarizes the explanation audits.

\subsection{Datasets and labels}

Models were trained on CODE100 and evaluated on held-out CODE-Test~\cite{ribeiro2020} and, as an external dataset, on PTB-XL~\cite{wagner2020ptbxl}. For PTB-XL, we mapped the diagnostic labels to the same six targets using strict SCP-ECG code presence. A domain specialist reviewed this mapping and the corresponding clinical definitions. The exact codes and class prevalence appear in Supplementary Note~1 and Supplementary Table~1.

\subsection{Turning points retain morphology and temporal coverage}

Graph construction begins by preprocessing each \SI{10}{\second} 12-lead record. We apply a fourth-order zero-phase Butterworth band-pass filter (\SIrange{0.5}{40}{\hertz}), resample to \SI{300}{\hertz} when needed, and normalize amplitude using the median across leads of their
95th-percentile amplitudes. A polarity-aware NeuroKit2 detector~\cite{makowski2021neurokit2} identifies R peaks independently in each lead. Consecutive R peaks define beat boundaries at their temporal midpoints, while the first and last beats extend to the signal boundaries. The quality-control criteria are reported in Supplementary Note~2; only \num{0.218}\% of exams are removed during validation of the full CODE graph build.

Within each beat, we use a low-pass signal to locate local extrema. These extrema, together with the detected R peak and the two beat endpoints, form the initial set of samples. We then limit the temporal gap between consecutive nodes to \SI{80}{\milli\second}. If two selected points are more than \SI{80}{\milli\second} apart, we insert support nodes at uniformly spaced samples until no gap exceeds this limit. Every node, including each support node, takes its amplitude from the preprocessed signal at its own sample, without interpolation, and retains its normalized within-beat time.

Natural visibility defines the connections between the retained nodes~\cite{lacasa2008}.
We number the retained nodes of a beat in time order, so node \(i\) has amplitude
\(a_i\) and time \(t_i\), in seconds. For nodes \(i<j\), an edge is added when every
intermediate retained node lies below the line joining them:
\begin{equation}
  a_k < a_i + \frac{a_j-a_i}{j-i}(k-i),
  \qquad i<k<j.
  \label{eq:visibility}
\end{equation}
Here \(i\), \(j\), and \(k\) are node positions, not times: the topology depends on
the order of the retained nodes but not on their spacing in time.
Each visible pair is oriented forward in time and stored as \(i\!\to\!j\). Each node stores amplitude and normalized within-beat time, \([a,t_{\mathrm{norm}}]\), while each edge stores \([\Delta t,\Delta a]\), with \(\Delta t=t_j-t_i>0\) and \(\Delta a=a_j-a_i\). Node times are in seconds, so \(\Delta t\) keeps absolute durations, whereas \(t_{\mathrm{norm}}\) is relative to the beat. Because each beat spans the two half R--R intervals around its R peak, its duration also encodes heart rate. The topology captures geometric visibility, while the node and edge features preserve amplitude and timing information.

\subsection{A hierarchical GNN follows ECG structure}

LYNX-ECG follows the beat \(\rightarrow\) lead \(\rightarrow\) exam hierarchy. At the beat level, a three-layer edge-aware Graph Isomorphism Network (GINE) encoder~\cite{xu2019gin,hu2020gine}, with hidden width 128 and multilayer perceptron (MLP) width 256, maps each directed beat graph to one embedding. Because within-beat edges point forward in time, later nodes can aggregate information from earlier nodes before global mean pooling. The resulting beat embeddings provide the inputs to the lead-level graph.

The beat representations from each lead form the next level of the hierarchy. A two-layer, four-head Graph Attention Network v2 (GATv2)~\cite{brody2022gatv2} exchanges information between temporally adjacent beats. Adjacent beats are connected in both directions, and learned direction embeddings distinguish forward from backward links. Learned lead embeddings encode lead identity. Gated attention~\cite{li2016gated} then combines the beat representations into one embedding for each lead.

At the exam level, softmax attention pools the available lead representations while masking absent leads. The primary model contains no explicit cross-lead graph edges, so information from different leads combines only during this final pooling step. A two-layer MLP head maps the resulting exam representation to logits for first-degree atrioventricular block (1dAVb), right and left bundle-branch block (RBBB/LBBB), sinus bradycardia (SB), atrial fibrillation (AF), and sinus tachycardia (ST).

\subsection{Dynamic sampling controls training composition}

Most CODE exams are negative for all six targets and would dominate batches
drawn from the full training distribution. Let \(D_{\mathrm{pos}}\) contain
every exam positive for at least one target and \(D_{\mathrm{neg}}\) contain
all-negative exams. At each epoch, we minimize binary cross-entropy over
\[
D_{\mathrm{pos}}
\cup
\operatorname{Sample}(D_{\mathrm{neg}},n_{\mathrm{neg}}),
\]
where \(n_{\mathrm{neg}}\) is the number of negative exams per epoch. We selected
\(n_{\mathrm{neg}}=\num{220514}\) together with the other training hyperparameters by
maximizing macro F1 on the CODE100 validation split (Supplementary Table~15). We draw the negative exams without replacement and select a new sample at the
start of each epoch. This procedure controls training composition while
exposing the model to different negative exams across epochs. Validation and test retain their original distributions. We select checkpoints
by validation macro F1 on a \num{2}\% patient-hash split disjoint from
CODE-Test. Every class and dataset uses a fixed decision threshold of
\num{0.5}. Supplementary Note~2 gives the full protocol.

\subsection{Baselines}

The deep neural network (DNN) baseline is the residual convolutional network
of Ribeiro et al., which processes the ECG time series~\cite{ribeiro2020}.
ECGFounder is an ECG foundation model developed through large-scale
pretraining~\cite{li2024ecgfounder}. We evaluated its architecture from random
initialization (ECGFounder-Scratch) and by fine-tuning pretrained weights
(ECGFounder-Finetuned).

Vision Transformer (ViT) uses attention over input
patches~\cite{dosovitskiy2021vit}, whereas EfficientNet uses a convolutional
architecture developed for efficient image
classification~\cite{tan2019efficientnet}. Both models take images, so we
rendered each ECG as a 12-lead plot. We removed baseline wander with a
zero-phase elliptic high-pass filter (0.8\,Hz cutoff) and cropped a 5\,s
window aligned to a detected QRS complex. Each plot shows the limb and
precordial leads in two columns on a standard ECG grid, as a
$1000\times720$-pixel image. We resized the images to the input size of each
model ($384\times384$ for ViT-B/16 and $300\times300$ for EfficientNet-B3) and
fine-tuned both models from ImageNet-pretrained weights. Supplementary Note~2
gives the remaining settings, and Supplementary Note~10 gives the paired comparisons.

\subsection{Representation and graph-design comparisons}

The node-selection comparison changes only the node selector. We kept the graph-construction rule, model, and
training protocol unchanged. Beat boundaries and effective batch size were
also matched. The two sparse selectors often covered similar temporal
neighborhoods, although turning points retained more nodes per beat on
average. This comparison therefore evaluates the selection strategies
without imposing an identical node budget.

The topology comparison keeps the turning-point nodes. For retained
nodes ordered in time as \(v_1,\ldots,v_n\), the temporal chain contains only
consecutive edges, \((v_i,v_{i+1})\). Natural visibility can instead connect
nonconsecutive nodes \(v_i\) and \(v_j\) when every intermediate node lies
below the straight line joining their amplitudes (equation~\eqref{eq:visibility}).
The chain therefore preserves only immediate temporal adjacency, whereas
visibility can introduce longer-range connections according to waveform
geometry. All four conditions of the \(2\times2\) design used the
same architecture and training protocol.

In the forward-only graph, an edge
\(v_i\rightarrow v_j\), with \(t_i<t_j\), carries information from an earlier
sample to a later one within the same beat. The bidirectional variant adds
the reverse edge, \(v_j\rightarrow v_i\), so the later sample can also send
information back to the earlier one. We kept the turning-point nodes,
visibility criterion, and edge-attribute definition fixed. Both variants were retrained with the same architecture and training protocol to isolate the effect of these additional reverse messages. We retained forward-only propagation because the reverse messages increased message passing without an established improvement in these comparisons.

\subsection{Regional explanation protocol}\label{sec:method-explain}

Each graph node retains its original timestamp, allowing node-level explanations and graph interventions to be mapped to ECG regions. We evaluate regional evidence with Integrated Gradients (IG)~\cite{sundararajan2017ig} and structural deletion.

For IG, we attribute the selected class logit to node features using a 64-step path from a reference input to the observed graph. We evaluate two references: an all-zero input and per-feature medians estimated from one million training-graph nodes. This reference provides the starting feature values for the attribution calculation. Feature attributions are summed within each node and converted to absolute node-level attribution, which measures attribution regardless of whether the contributions were positive or negative. Regional attribution is then aggregated relative to the R peak using PR (\(-0.35\) to \(-0.07\)\,s) for 1dAVb and QRS
(\(-0.06\) to \(+0.12\)\,s) for RBBB and LBBB. For each region, the lift divides its fraction of the total absolute attribution by its fraction of all nodes. It equals 1 when attribution is proportional to the node count, and values above 1 indicate more attribution per node than the overall average.

Structural deletion instead uses PR, QRS, and T regions delineated separately for each beat. A NeuroKit2 prominence delineator~\cite{makowski2021neurokit2} identifies P onset, QRS onset and offset, and T onset and offset from the stored R anchors. Its parameters are selected without using model outputs (Supplementary Note~8). The primary analysis includes lead--beat units for which all five landmarks are present and ordered, covering \num{89.1}\% of units. Deleting a region removes its nodes and incident edges and then links the last remaining node before the region to the first remaining node after it, so that each beat graph stays connected. Like every other edge, this edge carries the time and amplitude differences between its two nodes. The exam is then rescored.

The three regions define \(2^3=8\) deletion states, including the graph without deletions, which measures each region's effect both alone and after other regions have been removed. Exact Shapley values decompose the intact-minus-all-regions-deleted logit difference among PR, QRS, and T; each value averages a region's effect across all possible removal orders, and the region with the largest contribution is the leading region. The analysis is repeated across all three seeds. Fixed-window regions and a per-region availability rule provide sensitivity analyses; the latter is detailed in Supplementary Note~9.

The primary analysis uses only lead--beat units in which PR, QRS, and T are
all identified. In LBBB, excluded units had a longer median QRS duration
(\SI{173}{\milli\second} versus \SI{163}{\milli\second} for included units).
To test the effect of this exclusion, we repeated the audit with each available
region analyzed independently. Among correct LBBB verdicts, the fraction in
which QRS had the largest contribution increased from \num{94}\% to
\num{99}\%. RBBB gained one verdict, and 1dAVb changed only slightly (Supplementary Note~9). The audit depends both on the intervention and on the regions that can be reliably delineated.

\subsection{Timing and lead interventions}\label{sec:method-timing}

The timing tests change a trained model's input, not the model. We edit the beat graphs of an exam, reassemble the exam, and classify it again with the same weights. An exam's heart rate is 60 divided by the median R--R interval, computed per lead and then taken as the median across leads. The sequence tests edit beats in three ways. \emph{Shuffled} randomly reorders the beats within each lead. \emph{Template} replaces every beat of a lead with copies of its median-duration beat, so all beats become identical. \emph{Equalized} stretches or compresses each interior beat to the lead's median beat duration. Stretching a beat multiplies the time between its nodes (\(\Delta t\)) and keeps each node's relative position in the beat (\(t_{\mathrm{norm}}\)).

The uniform rate change stretches or compresses every beat by the ratio of the measured to the imposed heart rate, so the PR, QRS, and QT durations change with it. The interval-preserving change keeps each beat's segment from P onset to T offset intact and stretches or compresses only the time outside it. Each changed beat is rebuilt with the dataset rules. A beat that would keep less than 20 ms outside this segment keeps its original shape. An imposed rate counts as infeasible when more than 10\% of interior beats keep their shape, which is why fast imposed rates are often infeasible.

Since each lead embedding is computed independently, removing a lead removes only its embedding from the final pooling. We can therefore score all \(2^{12}\) subsets of leads from embeddings computed once per exam, with the empty subset giving a zero exam embedding. Shapley values then split the difference between the full-exam and empty-exam logits exactly among the 12 leads. The subset logits match the logits obtained by deleting the lead graphs from the input. For the timing and lead tests, we report the fraction of verdicts that keep or change a call. A verdict is the call of one trained model on one exam and label, and we pool verdicts over exams and the three seeds. The 95\% intervals of the timing tests (Supplementary Note~13 and Supplementary Tables~12 and~13) come from 2000 bootstrap resamples of exams, each resample keeping all three seeds of an exam.

\subsection{Use of LLMs}

We used large language models (LLMs) for generating initial prototype code, suggesting code optimizations, and proofreading and editing, always under full human supervision. The authors take full responsibility for the content of this article.

\subsection{Ethics statement}

This study complies with all relevant ethical regulations. Use of the CODE
data was approved by the Research Ethics Committee of the Universidade Federal
de Minas Gerais (protocol 49368496317.7.0000.5149)~\cite{ribeiro2020}. The
study is a secondary analysis of anonymized ECG exams stored by the Telehealth
Network of Minas Gerais, so the committee waived informed consent. Researchers
with access to CODE signed terms of confidentiality and data use. PTB-XL is an
openly available, anonymized dataset whose publication was approved by the
institutional ethics committee of the Physikalisch-Technische Bundesanstalt
(PTB-2020-1)~\cite{wagner2020ptbxl}.

\backmatter

\bmhead{Data availability}
PTB-XL~\cite{wagner2020ptbxl,wagner2022ptbxl103} and
CODE-Test~\cite{ribeiro2020,ribeiro2020codetest} are openly available. CODE100 is the training portion of the CODE
dataset~\cite{ribeiro2021codedataset}, which contains 12-lead ECG exams recorded
by the Telehealth Network of Minas Gerais and is under restricted access to
protect patient privacy. Researchers affiliated with educational or research
institutions can request access through the dataset's SciLifeLab Data
Repository record~\cite{ribeiro2021codedataset} by providing the name of
the principal investigator and host organization, contact details, and the
scientific purpose of the request. The Telehealth Network of Minas Gerais
evaluates each request individually, data use is restricted to non-commercial
research, and access is granted after a data use agreement is
signed~\cite{ribeiro2020}. A subset with 15\% of the patients, CODE-15\%, is
openly available~\cite{ribeiro2021code15}.

\bmhead{Code availability}
Our code is released at \url{https://github.com/aaaxn/LYNX-ECG} under the MIT License.

\bmhead{Acknowledgements}
SP acknowledges support from NSF grant \#2112471 (AI-EDGE) and a fellowship from the Translational Data Analytics Institute (TDAI) at Ohio State University. SP and WM also acknowledge the Translational Data Analytics Institute's Joint Human AI Systems (JHAS) theme for partially sponsoring WM's visit to Ohio State University where this collaboration was initiated. WM acknowledges support from the Brazilian agencies CNPq, CAPES and FAPEMIG, and projects CIIA-Saúde, IAIA-INCT on AI, INCTTildIAR, NIAR-Saúde, and the Brazilian Center on Algorithm Transparency and AI Safety.
ALPR is supported in part by Brazilian agencies CNPq, FAPEMIG and FAPESP, projects CIIA-Saúde and IATS-CARE, and SEIDIGI -- Brazilian Ministry of Health.
This work represents the authors' views and findings and does not necessarily reflect those of their funding agencies or their institutions.

\bmhead{Author contributions}
Conceptualization of initial ideas and method design: AX, WM, and SP during a collaboration visit. Refinement of ideas and methodology: all authors. Implementation: AX. Experimental design: all authors. Analysis of experimental results: all authors. Clinical interpretation and discussion of results: ALPR with all authors. Initial draft: AX. Draft refinement: all authors. Funding for the study: WM, SP, and MZ.

\bmhead{Competing interests}
The authors declare no competing interests.

\bmhead{Supplementary information}
The Supplementary Information contains Supplementary Notes~1--15, Supplementary Figs.~1--5, and Supplementary Tables~1--15.

\bibliography{refs}

\begin{thebibliography}{10}
\expandafter\ifx\csname url\endcsname\relax
  \def\url#1{\burl{#1}}\fi
\expandafter\ifx\csname urlprefix\endcsname\relax\def\urlprefix{URL }\fi
\providecommand{\bibinfo}[2]{#2}
\providecommand{\eprint}[2][]{\url{#2}}
\providecommand{\doi}[1]{\url{https://doi.org/#1}}
\bibcommenthead

\bibitem{ribeiro2020}
\bibinfo{author}{Ribeiro, A.~H.} \emph{et~al.}
\newblock \bibinfo{title}{Automatic diagnosis of the short-duration 12-lead {ECG} using a deep neural network}.
\newblock \emph{\bibinfo{journal}{Nature Communications}} \textbf{\bibinfo{volume}{11}}, \bibinfo{pages}{1760} (\bibinfo{year}{2020}).

\bibitem{wagner2020ptbxl}
\bibinfo{author}{Wagner, P.} \emph{et~al.}
\newblock \bibinfo{title}{{PTB-XL}, a large publicly available electrocardiography dataset}.
\newblock \emph{\bibinfo{journal}{Scientific Data}} \textbf{\bibinfo{volume}{7}}, \bibinfo{pages}{154} (\bibinfo{year}{2020}).

\bibitem{strodthoff2021ptbxlbench}
\bibinfo{author}{Strodthoff, N.}, \bibinfo{author}{Wagner, P.}, \bibinfo{author}{Schaeffter, T.} \& \bibinfo{author}{Samek, W.}
\newblock \bibinfo{title}{Deep learning for {ECG} analysis: Benchmarks and insights from {PTB-XL}}.
\newblock \emph{\bibinfo{journal}{IEEE Journal of Biomedical and Health Informatics}} \textbf{\bibinfo{volume}{25}}, \bibinfo{pages}{1519--1528} (\bibinfo{year}{2021}).

\bibitem{li2024ecgfounder}
\bibinfo{author}{Li, J.} \emph{et~al.}
\newblock \bibinfo{title}{An electrocardiogram foundation model built on over 10 million recordings with external evaluation across multiple domains}.
\newblock \emph{\bibinfo{journal}{arXiv preprint arXiv:2410.04133}}  (\bibinfo{year}{2024}).

\bibitem{wang2025melp}
\bibinfo{author}{Wang, F.}, \bibinfo{author}{Xu, J.} \& \bibinfo{author}{Yu, L.} \bibinfo{title}{From token to rhythm: A multi-scale approach for {ECG}-language pretraining}.
\newblock In \emph{\bibinfo{booktitle}{Proceedings of the 42nd International Conference on Machine Learning}}, Vol. \bibinfo{volume}{267} of \emph{\bibinfo{series}{Proceedings of Machine Learning Research}}, \bibinfo{pages}{65059--65074} (\bibinfo{year}{2025}).
\newblock \urlprefix\url{https://proceedings.mlr.press/v267/wang25du.html}.

\bibitem{zhou2025htuning}
\bibinfo{author}{Zhou, R.}, \bibinfo{author}{Zhang, Y.} \& \bibinfo{author}{Dong, Y.} \bibinfo{title}{{H-Tuning}: Toward low-cost and efficient {ECG}-based cardiovascular disease detection with pre-trained models}.
\newblock In \emph{\bibinfo{booktitle}{Proceedings of the 42nd International Conference on Machine Learning}}, Vol. \bibinfo{volume}{267} of \emph{\bibinfo{series}{Proceedings of Machine Learning Research}}, \bibinfo{pages}{79548--79569} (\bibinfo{year}{2025}).
\newblock \urlprefix\url{https://proceedings.mlr.press/v267/zhou25aj.html}.

\bibitem{sundararajan2017ig}
\bibinfo{author}{Sundararajan, M.}, \bibinfo{author}{Taly, A.} \& \bibinfo{author}{Yan, Q.} \bibinfo{title}{Axiomatic attribution for deep networks}.
\newblock In \emph{\bibinfo{booktitle}{International Conference on Machine Learning (ICML)}} (\bibinfo{year}{2017}).

\bibitem{sturmfels2020baselines}
\bibinfo{author}{Sturmfels, P.}, \bibinfo{author}{Lundberg, S.} \& \bibinfo{author}{Lee, S.-I.}
\newblock \bibinfo{title}{Visualizing the impact of feature attribution baselines}.
\newblock \emph{\bibinfo{journal}{Distill}}  (\bibinfo{year}{2020}).
\newblock \bibinfo{note}{\url{https://distill.pub/2020/attribution-baselines/}}.

\bibitem{ancona2018towards}
\bibinfo{author}{Ancona, M.}, \bibinfo{author}{Ceolini, E.}, \bibinfo{author}{{\"O}ztireli, C.} \& \bibinfo{author}{Gross, M.} \bibinfo{title}{Towards better understanding of gradient-based attribution methods for deep neural networks}.
\newblock In \emph{\bibinfo{booktitle}{International Conference on Learning Representations (ICLR)}} (\bibinfo{year}{2018}).

\bibitem{yuan2022survey}
\bibinfo{author}{Yuan, H.}, \bibinfo{author}{Yu, H.}, \bibinfo{author}{Gui, S.} \& \bibinfo{author}{Ji, S.}
\newblock \bibinfo{title}{Explainability in graph neural networks: A taxonomic survey}.
\newblock \emph{\bibinfo{journal}{IEEE Transactions on Pattern Analysis and Machine Intelligence}} \textbf{\bibinfo{volume}{45}}, \bibinfo{pages}{5782--5799} (\bibinfo{year}{2022}).

\bibitem{ying2019gnnexplainer}
\bibinfo{author}{Ying, Z.}, \bibinfo{author}{Bourgeois, D.}, \bibinfo{author}{You, J.}, \bibinfo{author}{Zitnik, M.} \& \bibinfo{author}{Leskovec, J.} \bibinfo{title}{{GNNExplainer}: Generating explanations for graph neural networks}.
\newblock In \emph{\bibinfo{booktitle}{Advances in Neural Information Processing Systems (NeurIPS)}} (\bibinfo{year}{2019}).

\bibitem{luo2020pgexplainer}
\bibinfo{author}{Luo, D.} \emph{et~al.} \bibinfo{title}{Parameterized explainer for graph neural network}.
\newblock In \emph{\bibinfo{booktitle}{Advances in Neural Information Processing Systems (NeurIPS)}} (\bibinfo{year}{2020}).

\bibitem{yuan2021subgraphx}
\bibinfo{author}{Yuan, H.}, \bibinfo{author}{Yu, H.}, \bibinfo{author}{Wang, J.}, \bibinfo{author}{Li, K.} \& \bibinfo{author}{Ji, S.} \bibinfo{title}{On explainability of graph neural networks via subgraph explorations}.
\newblock In \emph{\bibinfo{booktitle}{International Conference on Machine Learning (ICML)}} (\bibinfo{year}{2021}).

\bibitem{azzolin2025formal}
\bibinfo{author}{Azzolin, S.}, \bibinfo{author}{Malhotra, S.}, \bibinfo{author}{Passerini, A.} \& \bibinfo{author}{Teso, S.} \bibinfo{title}{Beyond topological self-explainable {GNNs}: A formal explainability perspective}.
\newblock In \emph{\bibinfo{booktitle}{Proceedings of the 42nd International Conference on Machine Learning}}, Vol. \bibinfo{volume}{267} of \emph{\bibinfo{series}{Proceedings of Machine Learning Research}}, \bibinfo{pages}{2045--2080} (\bibinfo{year}{2025}).
\newblock \urlprefix\url{https://proceedings.mlr.press/v267/azzolin25a.html}.

\bibitem{tai2025redundancy}
\bibinfo{author}{Tai, W.}, \bibinfo{author}{Zhong, T.}, \bibinfo{author}{Trajcevski, G.} \& \bibinfo{author}{Zhou, F.} \bibinfo{title}{Redundancy undermines the trustworthiness of self-interpretable {GNNs}}.
\newblock In \emph{\bibinfo{booktitle}{Proceedings of the 42nd International Conference on Machine Learning}}, Vol. \bibinfo{volume}{267} of \emph{\bibinfo{series}{Proceedings of Machine Learning Research}}, \bibinfo{pages}{58169--58188} (\bibinfo{year}{2025}).
\newblock \urlprefix\url{https://proceedings.mlr.press/v267/tai25a.html}.

\bibitem{yang2025tif}
\bibinfo{author}{Yang, J.} \emph{et~al.} \bibinfo{title}{From {GNNs} to trees: Multi-granular interpretability for graph neural networks}.
\newblock In \emph{\bibinfo{booktitle}{International Conference on Learning Representations (ICLR)}} (\bibinfo{year}{2025}).

\bibitem{zhang2025confexplainer}
\bibinfo{author}{Zhang, J.}, \bibinfo{author}{Liu, X.}, \bibinfo{author}{Luo, D.} \& \bibinfo{author}{Wei, H.} \bibinfo{title}{Is your explanation reliable: Confidence-aware explanation on graph neural networks}.
\newblock In \emph{\bibinfo{booktitle}{Proceedings of the ACM SIGKDD Conference on Knowledge Discovery and Data Mining (KDD)}} (\bibinfo{year}{2025}).

\bibitem{lacasa2008}
\bibinfo{author}{Lacasa, L.}, \bibinfo{author}{Luque, B.}, \bibinfo{author}{Ballesteros, F.}, \bibinfo{author}{Luque, J.} \& \bibinfo{author}{Nu{\~n}o, J.~C.}
\newblock \bibinfo{title}{From time series to complex networks: The visibility graph}.
\newblock \emph{\bibinfo{journal}{Proceedings of the National Academy of Sciences}} \textbf{\bibinfo{volume}{105}}, \bibinfo{pages}{4972--4975} (\bibinfo{year}{2008}).

\bibitem{oliveira2024vgecg}
\bibinfo{author}{Oliveira, R.~F.}, \bibinfo{author}{Moreira, G. J.~P.}, \bibinfo{author}{Freitas, V. L.~S.} \& \bibinfo{author}{Luz, E. J.~S.}
\newblock \bibinfo{title}{Leveraging visibility graphs for enhanced arrhythmia classification with graph convolutional networks}.
\newblock \emph{\bibinfo{journal}{arXiv preprint arXiv:2404.15367}}  (\bibinfo{year}{2024}).

\bibitem{liu2025softshape}
\bibinfo{author}{Liu, Z.} \emph{et~al.} \bibinfo{title}{Learning soft sparse shapes for efficient time-series classification}.
\newblock In \emph{\bibinfo{booktitle}{Proceedings of the 42nd International Conference on Machine Learning}}, Vol. \bibinfo{volume}{267} of \emph{\bibinfo{series}{Proceedings of Machine Learning Research}}, \bibinfo{pages}{39032--39059} (\bibinfo{year}{2025}).
\newblock \urlprefix\url{https://proceedings.mlr.press/v267/liu25ar.html}.

\bibitem{zhang2026medmamba}
\bibinfo{author}{Zhang, D.} \emph{et~al.} \bibinfo{title}{{MedMamba}: Multi-view state space models with adaptive graph learning for medical time series classification}.
\newblock In \emph{\bibinfo{booktitle}{Proceedings of the 43rd International Conference on Machine Learning}} (\bibinfo{year}{2026}).
\newblock \urlprefix\url{https://icml.cc/virtual/2026/poster/61414}.

\bibitem{lee2025ecggraphnet}
\bibinfo{author}{Lee, M.}, \bibinfo{author}{Lim, J.} \& \bibinfo{author}{Kim, J.}
\newblock \bibinfo{title}{{ECG-GraphNet}: Advanced arrhythmia classification based on graph convolutional networks}.
\newblock \emph{\bibinfo{journal}{Heart Rhythm O2}} \textbf{\bibinfo{volume}{6}}, \bibinfo{pages}{1199--1211} (\bibinfo{year}{2025}).

\bibitem{maurer2026xgnn4mi}
\bibinfo{author}{Maurer, M.~C.} \emph{et~al.}
\newblock \bibinfo{title}{{xGNN4MI}: Explainability of graph neural networks in 12-lead electrocardiography for cardiovascular disease classification}.
\newblock \emph{\bibinfo{journal}{npj Digital Medicine}} \textbf{\bibinfo{volume}{9}}, \bibinfo{pages}{256} (\bibinfo{year}{2026}).

\bibitem{satuluri2011local}
\bibinfo{author}{Satuluri, V.}, \bibinfo{author}{Parthasarathy, S.} \& \bibinfo{author}{Ruan, Y.} \bibinfo{title}{Local graph sparsification for scalable clustering}.
\newblock In \emph{\bibinfo{booktitle}{Proceedings of the 2011 ACM SIGMOD International Conference on Management of data}}, \bibinfo{pages}{721--732} (\bibinfo{year}{2011}).

\bibitem{dosovitskiy2021vit}
\bibinfo{author}{Dosovitskiy, A.} \emph{et~al.} \bibinfo{title}{An image is worth 16x16 words: Transformers for image recognition at scale}.
\newblock In \emph{\bibinfo{booktitle}{International Conference on Learning Representations (ICLR)}} (\bibinfo{year}{2021}).
\newblock \urlprefix\url{https://openreview.net/forum?id=YicbFdNTTy}.

\bibitem{tan2019efficientnet}
\bibinfo{author}{Tan, M.} \& \bibinfo{author}{Le, Q.} \bibinfo{title}{{EfficientNet}: Rethinking model scaling for convolutional neural networks}.
\newblock In \emph{\bibinfo{booktitle}{Proceedings of the 36th International Conference on Machine Learning}}, Vol.~\bibinfo{volume}{97} of \emph{\bibinfo{series}{Proceedings of Machine Learning Research}}, \bibinfo{pages}{6105--6114} (\bibinfo{publisher}{PMLR}, \bibinfo{year}{2019}).
\newblock \urlprefix\url{https://proceedings.mlr.press/v97/tan19a.html}.

\bibitem{surawicz2009}
\bibinfo{author}{Surawicz, B.}, \bibinfo{author}{Childers, R.}, \bibinfo{author}{Deal, B.~J.}, \bibinfo{author}{Gettes, L.~S.} \emph{et~al.}
\newblock \bibinfo{title}{{AHA/ACCF/HRS} recommendations for the standardization and interpretation of the electrocardiogram: Part {III}: Intraventricular conduction disturbances}.
\newblock \emph{\bibinfo{journal}{Circulation}} \textbf{\bibinfo{volume}{119}}, \bibinfo{pages}{e235--e240} (\bibinfo{year}{2009}).

\bibitem{shapley1953value}
\bibinfo{author}{Shapley, L.~S.} \bibinfo{title}{A value for $n$-person games}.
\newblock In \emph{\bibinfo{booktitle}{Contributions to the Theory of Games {II} (eds Kuhn, H. W. \& Tucker, A. W.)}}, no.~\bibinfo{number}{28} in \bibinfo{series}{Annals of Mathematics Studies}, \bibinfo{pages}{307--317} (\bibinfo{publisher}{Princeton University Press}, \bibinfo{year}{1953}).

\bibitem{vadlamani2026}
\bibinfo{author}{Vadlamani, A.~T.}, \bibinfo{author}{Chang, J.~H.}, \bibinfo{author}{Parthasarathy, S.} \& \bibinfo{author}{Paul, S.}
\newblock \bibinfo{title}{Transfer calibrated prediction powered inference}.
\newblock \emph{\bibinfo{journal}{arXiv preprint arXiv:2609.34156}}  (\bibinfo{year}{2026}).

\bibitem{makowski2021neurokit2}
\bibinfo{author}{Makowski, D.} \emph{et~al.}
\newblock \bibinfo{title}{{NeuroKit2}: A python toolbox for neurophysiological signal processing}.
\newblock \emph{\bibinfo{journal}{Behavior Research Methods}} \textbf{\bibinfo{volume}{53}}, \bibinfo{pages}{1689--1696} (\bibinfo{year}{2021}).

\bibitem{xu2019gin}
\bibinfo{author}{Xu, K.}, \bibinfo{author}{Hu, W.}, \bibinfo{author}{Leskovec, J.} \& \bibinfo{author}{Jegelka, S.} \bibinfo{title}{How powerful are graph neural networks?}
\newblock In \emph{\bibinfo{booktitle}{International Conference on Learning Representations (ICLR)}} (\bibinfo{year}{2019}).

\bibitem{hu2020gine}
\bibinfo{author}{Hu, W.} \emph{et~al.} \bibinfo{title}{Strategies for pre-training graph neural networks}.
\newblock In \emph{\bibinfo{booktitle}{International Conference on Learning Representations (ICLR)}} (\bibinfo{year}{2020}).

\bibitem{brody2022gatv2}
\bibinfo{author}{Brody, S.}, \bibinfo{author}{Alon, U.} \& \bibinfo{author}{Yahav, E.} \bibinfo{title}{How attentive are graph attention networks?}
\newblock In \emph{\bibinfo{booktitle}{International Conference on Learning Representations (ICLR)}} (\bibinfo{year}{2022}).

\bibitem{li2016gated}
\bibinfo{author}{Li, Y.}, \bibinfo{author}{Tarlow, D.}, \bibinfo{author}{Brockschmidt, M.} \& \bibinfo{author}{Zemel, R.} \bibinfo{title}{Gated graph sequence neural networks}.
\newblock In \emph{\bibinfo{booktitle}{International Conference on Learning Representations (ICLR)}} (\bibinfo{year}{2016}).

\bibitem{wagner2022ptbxl103}
\bibinfo{author}{Wagner, P.}, \bibinfo{author}{Strodthoff, N.}, \bibinfo{author}{Bousseljot, R.-D.}, \bibinfo{author}{Samek, W.} \& \bibinfo{author}{Schaeffter, T.}
\newblock \bibinfo{title}{{PTB-XL}, a large publicly available electrocardiography dataset (version 1.0.3)}.
\newblock \bibinfo{howpublished}{PhysioNet} (\bibinfo{year}{2022}).
\newblock \urlprefix\url{https://doi.org/10.13026/kfzx-aw45}.

\bibitem{ribeiro2020codetest}
\bibinfo{author}{Ribeiro, A.~H.} \emph{et~al.}
\newblock \bibinfo{title}{{CODE}-test: An annotated 12-lead {ECG} dataset (version 1.0.3)}.
\newblock \bibinfo{howpublished}{Zenodo} (\bibinfo{year}{2020}).
\newblock \urlprefix\url{https://doi.org/10.5281/zenodo.3625006}.

\bibitem{ribeiro2021codedataset}
\bibinfo{author}{Ribeiro, A.~H.} \emph{et~al.}
\newblock \bibinfo{title}{{CODE} dataset}.
\newblock \bibinfo{howpublished}{SciLifeLab Data Repository} (\bibinfo{year}{2021}).
\newblock \urlprefix\url{https://doi.org/10.17044/scilifelab.15169716}.

\bibitem{ribeiro2021code15}
\bibinfo{author}{Ribeiro, A.~H.} \emph{et~al.}
\newblock \bibinfo{title}{{CODE}-15\%: a large scale annotated dataset of 12-lead {ECGs} (version 1.0.0)}.
\newblock \bibinfo{howpublished}{Zenodo} (\bibinfo{year}{2021}).
\newblock \urlprefix\url{https://doi.org/10.5281/zenodo.4916206}.

\end{thebibliography}

\clearpage
\setcounter{table}{0}
\setcounter{figure}{0}
\renewcommand{\tablename}{Extended Data Table}
\renewcommand{\figurename}{Extended Data Fig.}
\renewcommand{\theHtable}{ED.\arabic{table}}
\renewcommand{\theHfigure}{ED.\arabic{figure}}

\begin{sidewaystable}[p]
  \caption{Per-class F1. Values are mean $\pm$ sample SD across three seeds.
  Boldface marks the highest mean and underline marks the second-highest mean (ties within 0.001). The macro column repeats
  Table~\ref{tab:performance}. Supplementary Table~5
  gives the per-class AUROC.}
  \label{tab:perclass-f1}
  \centering
  \footnotesize
  \setlength{\tabcolsep}{4pt}
  \begin{tabular}{@{}lccccccc@{}}
    \toprule
    Model & 1dAVb & RBBB & LBBB & SB & AF & ST & Macro \\
    \midrule
    \multicolumn{8}{@{}l}{\textit{CODE-Test ($n=827$)}} \\
    LYNX-ECG & $\underline{0.783\pm0.076}$ & $0.920\pm0.037$ & $0.959\pm0.010$ & $\mathbf{0.843\pm0.037}$ & $\mathbf{0.918\pm0.018}$ & $0.911\pm0.014$ & $\mathbf{0.889\pm0.004}$ \\
    DNN~\cite{ribeiro2020} & $0.717\pm0.104$ & $0.913\pm0.020$ & $\underline{0.966\pm0.000}$ & $0.804\pm0.025$ & $0.773\pm0.040$ & $\underline{0.922\pm0.013}$ & $0.849\pm0.023$ \\
    ViT~\cite{dosovitskiy2021vit} & $0.724\pm0.052$ & $\underline{0.926\pm0.008}$ & $\mathbf{0.983\pm0.000}$ & $0.783\pm0.016$ & $0.848\pm0.016$ & $\mathbf{0.923\pm0.012}$ & $0.864\pm0.011$ \\
    EfficientNet~\cite{tan2019efficientnet} & $\mathbf{0.800\pm0.021}$ & $\mathbf{0.948\pm0.009}$ & $\underline{0.966\pm0.017}$ & $0.758\pm0.013$ & $0.860\pm0.063$ & $0.901\pm0.007$ & $\underline{0.872\pm0.009}$ \\
    ECGFounder-Scratch~\cite{li2024ecgfounder} & $0.743\pm0.043$ & $0.901\pm0.052$ & $0.960\pm0.009$ & $\underline{0.811\pm0.031}$ & $\underline{0.867\pm0.029}$ & $0.898\pm0.013$ & $0.863\pm0.004$ \\
    ECGFounder-Finetuned~\cite{li2024ecgfounder} & $0.756\pm0.059$ & $0.923\pm0.010$ & $0.897\pm0.058$ & $0.796\pm0.043$ & $0.852\pm0.071$ & $0.921\pm0.012$ & $0.858\pm0.035$ \\
    \midrule
    \multicolumn{8}{@{}l}{\textit{PTB-XL ($n=\num{21799}$)}} \\
    LYNX-ECG & $0.441\pm0.018$ & $\underline{0.773\pm0.029}$ & $0.734\pm0.036$ & $\mathbf{0.697\pm0.009}$ & $\underline{0.920\pm0.002}$ & $0.815\pm0.021$ & $\underline{0.730\pm0.001}$ \\
    DNN~\cite{ribeiro2020} & $0.458\pm0.022$ & $\mathbf{0.801\pm0.003}$ & $0.749\pm0.022$ & $0.604\pm0.009$ & $0.884\pm0.019$ & $0.814\pm0.011$ & $0.718\pm0.003$ \\
    ViT~\cite{dosovitskiy2021vit} & $0.459\pm0.013$ & $0.745\pm0.015$ & $0.712\pm0.045$ & $0.657\pm0.011$ & $0.909\pm0.003$ & $0.808\pm0.008$ & $0.715\pm0.010$ \\
    EfficientNet~\cite{tan2019efficientnet} & $0.458\pm0.025$ & $0.761\pm0.018$ & $\mathbf{0.777\pm0.014}$ & $0.647\pm0.013$ & $0.916\pm0.007$ & $0.806\pm0.021$ & $0.728\pm0.003$ \\
    ECGFounder-Scratch~\cite{li2024ecgfounder} & $\mathbf{0.482\pm0.033}$ & $0.751\pm0.027$ & $0.725\pm0.041$ & $0.672\pm0.009$ & $\mathbf{0.924\pm0.006}$ & $\mathbf{0.828\pm0.022}$ & $\underline{0.730\pm0.013}$ \\
    ECGFounder-Finetuned~\cite{li2024ecgfounder} & $\underline{0.480\pm0.005}$ & $0.739\pm0.003$ & $\underline{0.761\pm0.016}$ & $\underline{0.684\pm0.008}$ & $0.915\pm0.005$ & $\underline{0.827\pm0.005}$ & $\mathbf{0.734\pm0.003}$ \\
    \bottomrule
  \end{tabular}
\end{sidewaystable}

\begin{table}[p]
  \caption{Graph topology and edge attributes on CODE-Test. Derived edge
    attributes encode the time and amplitude differences between connected nodes,
    \((\Delta t,\Delta a)\). Values are mean \(\pm\) SD across three matched
    seeds; contrast rows report paired within-seed differences in percentage
    points. The interaction row is the visibility--chain contrast with the
    derived edge attributes minus the same contrast without them. Lower expected calibration error (ECE) is better.}
  \label{tab:topology-factorial}
  \centering
  \footnotesize
  \setlength{\tabcolsep}{4pt}
  \begin{tabular}{lccc}
    \toprule
    Configuration or contrast & Macro F1@0.5 & Macro AUROC & Macro ECE \\
    \midrule
    Visibility + edge attributes (LYNX-ECG)
      & \(\mathbf{0.889\pm0.004}\)
      & \(\mathbf{0.99842\pm0.00005}\)
      & \(\mathbf{0.00799\pm0.00102}\) \\
    Visibility + no edge attributes
      & \(0.799\pm0.061\)
      & \(0.99530\pm0.00199\)
      & \(0.00983\pm0.00173\) \\
    Temporal chain + edge attributes
      & \(0.863\pm0.010\)
      & \(0.99818\pm0.00013\)
      & \(0.00882\pm0.00085\) \\
    Temporal chain + no edge attributes
      & \(0.845\pm0.024\)
      & \(0.99721\pm0.00025\)
      & \(0.00912\pm0.00064\) \\
    \midrule
    Visibility--chain contrast, with attributes
      & \(+2.640\pm1.105\)
      & \(+0.023\pm0.016\)
      & \(-0.083\pm0.116\) \\
    Interaction
      & \(+7.281\pm2.740\)
      & \(+0.214\pm0.163\)
      & \(-0.154\pm0.211\) \\
    \bottomrule
  \end{tabular}
\end{table}

\clearpage

\begin{table}[p]
  \caption{Regional IG attribution under two reference inputs.
  The analysis includes 85 label-positive CODE-Test exams,
  comprising 92 positive exam--class pairs. Ranges span the minimum
  and maximum of the three per-seed median lifts. A lift of 1
  indicates attribution proportional to the region's node count.
  The comparison region is QRS for 1dAVb and PR for RBBB and LBBB.
  The feature-median baseline uses training-set feature medians.}
  \label{tab:alignment}
  \centering
  \small
  \begin{tabular}{llcccc}
    \toprule
    & & \multicolumn{2}{c}{Zero baseline}
      & \multicolumn{2}{c}{Feature-median baseline} \\
    \cmidrule(lr){3-4}\cmidrule(l){5-6}
    Class & Expected region & Expected & Comparison
      & Expected & Comparison \\
    \midrule
    1dAVb & PR
      & 0.753--0.780 & 1.980--2.109
      & 1.096--1.257 & 1.239--1.312 \\
    RBBB & QRS
      & 2.156--2.208 & 0.499--0.592
      & 1.469--1.764 & 0.928--1.053 \\
    LBBB & QRS
      & 2.048--2.113 & 0.682--0.742
      & 1.335--1.514 & 0.996--1.072 \\
    \bottomrule
  \end{tabular}
\end{table}

\begin{table}[p]
  \caption{Effect of region definition on structural-deletion verdicts.
  All three definitions are evaluated on the same available beats.
  Cells report how often the expected region leads among correctly
  classified verdicts. ``T leads'' counts 1dAVb verdicts in which the
  T region has the largest Shapley contribution. Both fixed-window
  definitions use PR from $-0.35$ to $-0.07$\,s and QRS from $-0.06$ to
  $+0.12$\,s relative to the R peak and differ only in the T window: the
  class-specific variant sets a different T window for each class, and the standardized window spans
  $+0.15$ to $+0.45$\,s for every class (Supplementary Note~8).}
  \label{tab:region-ablation}
  \centering
  \small
  \begin{tabular}{lccccc}
    \toprule
    Region definition & 1dAVb & RBBB & LBBB & All & T leads \\
    \midrule
    Per-beat delineation & 47/59 & 95/96 & 78/83 & 220/238 (92\%) & 1 \\
    Fixed, class-specific & 48/59 & 89/96 & 68/83 & 205/238 (86\%) & 5 \\
    Fixed, standardized  & 49/59 & 89/96 & 67/83 & 205/238 (86\%) & 5 \\
    \bottomrule
  \end{tabular}
\end{table}

\clearpage

\begin{figure}[p]
  \centering
  \includegraphics[width=\textwidth]{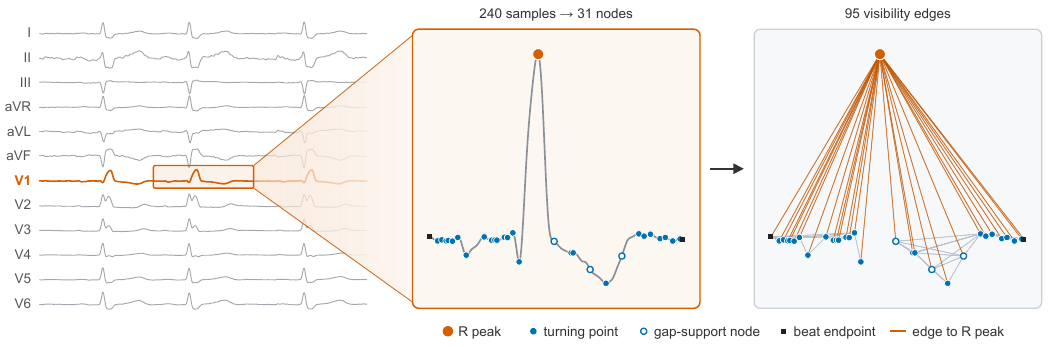}
  \caption{\textbf{Each beat becomes a sparse graph that keeps its morphology.} One V1 beat of a CODE-Test exam keeps 31 of its 240 samples (\num{87}\% fewer). Natural visibility connects these nodes with 95 edges; R-peak edges are highlighted.}
  \label{fig:construction}
\end{figure}

\end{document}


\maketitle

\setcounter{tocdepth}{1}
\tableofcontents
\clearpage

\section{Datasets and label protocol}
\label{app:datasets}

We train on the full CODE100 dataset~\cite{ribeiro2020}, accessed with
authorization from the Telehealth Network of Minas Gerais, and evaluate on the
openly released CODE-Test cohort. PTB-XL v1.0.3
~\cite{wagner2020ptbxl,wagner2022ptbxl103,goldberger2000physionet} provides
\num{21799} ECGs from \num{18869} patients under CC BY 4.0. The model reads
CODE records at \SI{300}{\hertz}; the public CODE files contain
\SI{400}{\hertz} signals. PTB-XL provides original \SI{500}{\hertz} records
and \SI{100}{\hertz} downsampled records. Every graph artifact uses the same
preprocessing described below and in Methods. After graph
quality control, Supplementary Table~\ref{tab:prevalence} reports the number of exams and the
per-class positive counts in each cohort. Because the task is multilabel, its
class counts need not sum to the number of exams.

The six targets are first-degree atrioventricular block (1dAVb), complete right
and left bundle-branch block (RBBB and LBBB), sinus bradycardia (SB), atrial
fibrillation (AF), and sinus tachycardia (ST). CODE training labels combine
cardiologist reports with interval and rhythm checks, including PR
$\geq\!190$\,ms for 1dAVb, QRS $\geq\!115$\,ms for bundle-branch block, heart
rate $\leq\!50$\,bpm for SB, and heart rate $\geq\!100$\,bpm for
ST~\cite{ribeiro2020}. Two certified cardiologists independently read
CODE-Test, with senior adjudication of disagreements.

PTB-XL uses SCP-ECG ontology codes rather than the CODE interval and rhythm
rules. We map exact SCP-code presence to the six targets in
the last row of Supplementary Table~\ref{tab:prevalence}. A domain specialist reviewed the class definitions and
mapping. The mapping includes first-degree atrioventricular block, complete
bundle-branch blocks, SB, ST, and AF, while excluding atrial flutter. Thus,
PTB-XL is an external evaluation under a strict ontology-to-task mapping, not a
relabeling under the CODE annotation protocol.

\section{Matched preprocessing, model, and training protocol}
\label{app:hparams}

\paragraph*{Preprocessing and graph quality control.}
We apply a fourth-order zero-phase Butterworth band-pass filter
(\SIrange{0.5}{40}{\hertz}), resample to \SI{300}{\hertz}, and normalize by
the median across leads of each lead's 95th-percentile absolute amplitude. A
polarity-aware NeuroKit2 0.2.13 detector~\cite{makowski2021neurokit2} locates
R peaks in each lead. Midpoints between adjacent R peaks delimit beats, with
the first and last beats extended to the signal boundaries. A
\SI{40}{\hertz} smoothing pass supplies local extrema. The retained nodes are
these extrema, the R peak, both beat endpoints, and fixed-gap support nodes
that enforce a maximum \SI{80}{\milli\second} gap. Node amplitudes come from
the preprocessed, unsmoothed beat.

Each lead must have at least three ordered, in-range R peaks and a median heart
rate between 30 and 240 beats/min. Nonboundary beats must contain finite
values, last \SIrange{0.25}{4.0}{\second}, contain at least three samples, and
have amplitude range greater than $10^{-4}$. Boundary beats bypass the
duration check. A valid lead contains at least three valid beats, and a valid
exam contains at least one valid lead.

\paragraph*{Shared architecture.}
Runs initialized with different seeds use hidden width 128, a three-layer GINE
encoder with MLP width 256, and a two-layer, four-head GATv2 encoder. Node
features are $[a,t_{\mathrm{norm}}]$, edge features are
$[\Delta t,\Delta a]$, and learned signed step embeddings distinguish forward
and backward links between adjacent beats. Attentional intra-lead pooling,
softmax inter-lead pooling, and a two-layer MLP six-logit head complete the hierarchy.
LYNX-ECG contains \num{327371} trainable parameters.

\paragraph*{Training and model selection.}
Let $y_n\in\{0,1\}^{6}$ denote the multilabel target for exam $n$. We partition
the training data into
$D_{\mathrm{pos}}=\{n:\sum_i y_{n,i}>0\}$ and
$D_{\mathrm{neg}}=\{n:\sum_i y_{n,i}=0\}$, with \num{233959} and \num{2003166}
exams in CODE100 training (Supplementary Table~\ref{tab:prevalence}). Every epoch contains all
exams in $D_{\mathrm{pos}}$ and a without-replacement sample of $n_{\mathrm{neg}}$
exams from $D_{\mathrm{neg}}$; the negative sample is redrawn each epoch. We
selected $n_{\mathrm{neg}}=\num{220514}$ by hyperparameter search together with
the other training hyperparameters (Supplementary Table~\ref{tab:hpo}). Each epoch therefore
holds \num{454473} exams, close to one negative per positive, and covers about
\num{11}\% of $D_{\mathrm{neg}}$. Validation and test distributions are
unchanged. Checkpoints are selected by validation macro F1 on the disjoint
\num{2}\% patient-hash split. Unless stated otherwise, all threshold-dependent
results use a fixed \num{0.5} threshold and three independently initialized
runs. Supplementary Table~\ref{tab:hpo} gives the declared hyperparameter ranges and the
selected configuration shared by these runs.

All representation comparisons retain the CODE100 patient-hash split,
CODE-Test artifact, architecture, optimizer, checkpoint rule, three seed
identifiers, and fixed threshold. The $2\times2$ topology ablation crosses
stored visibility or a directed chain over consecutive turning points with
derived or zero edge attributes. The matched baselines use two RTX 5090
GPUs, while the original baseline study used eight RTX A6000 GPUs. The baselines runs
preserve the per-GPU batch size and therefore use a smaller total batch.
ECGFounder-Scratch uses fp32 because its normalization-free Net1D overflows
fp16 when trained from random initialization.

Versioned scripts produce the reported metrics and bind them to the dataset
and checkpoint artifacts. The CODE data-use agreement prevents redistribution
of the restricted waveforms.

\section{Per-class performance across datasets}
\label{app:roc}

Supplementary Table~\ref{tab:performance-code} consolidates the per-class results on
CODE-Test (panel A) and strict-SCP PTB-XL (panel B). The
per-class AUROC, F1, precision, and recall entries are ranges across the three
runs. Per-class AUPRC and every macro metric are means $\pm$ sample SD across
runs.

\begin{figure}[p]
  \centering
  \includegraphics[width=\linewidth,height=0.78\textheight,keepaspectratio]{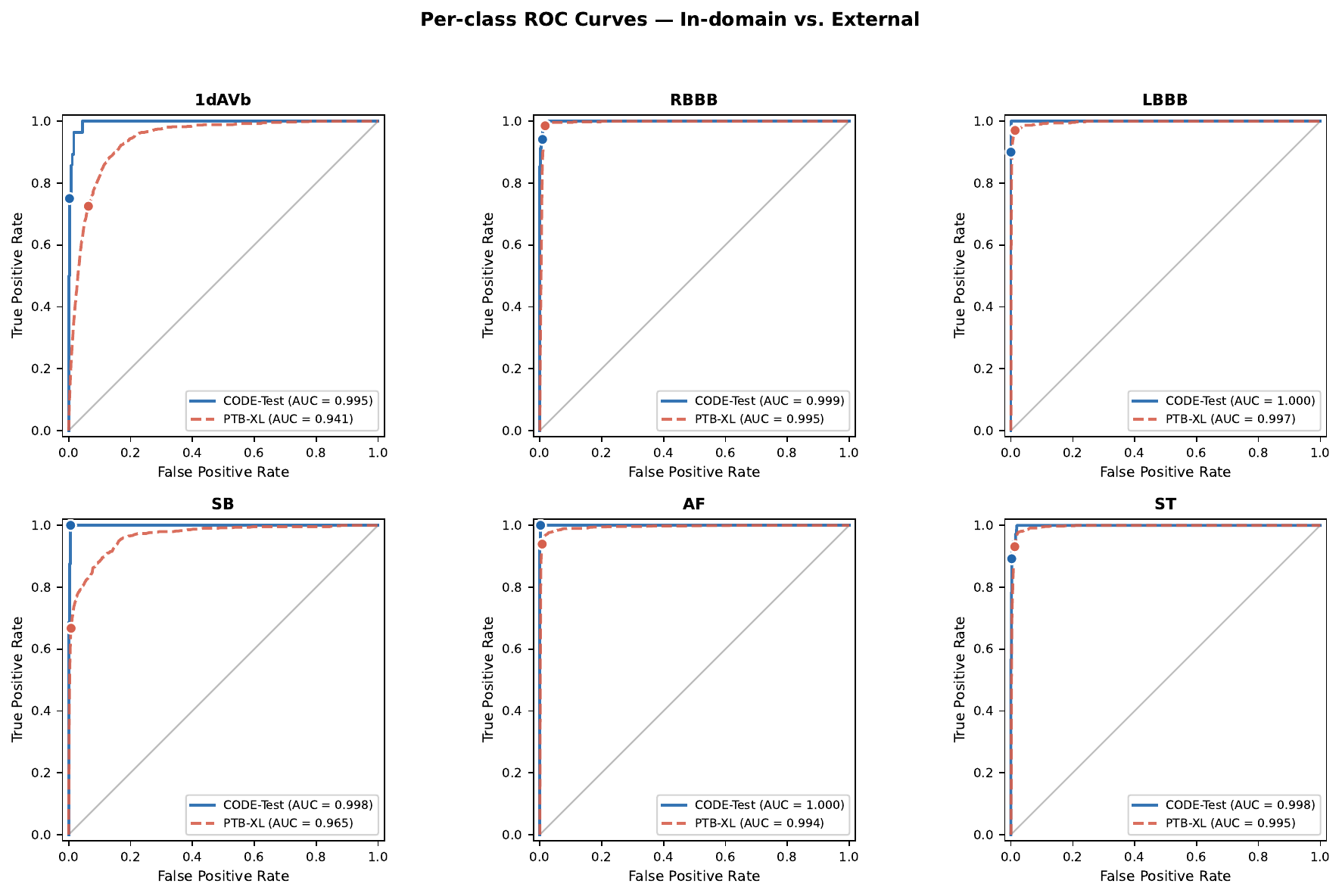}
  \caption{\textbf{External ranking degrades most for 1dAVb.} Run~A curves
  mark the fixed-\num{0.5} operating point;
  Supplementary Table~\ref{tab:performance-code} reports per-class results across all
  three seeds.}
    \label{fig:roc}
\end{figure}

Threshold-dependent metrics in this table use the fixed \num{0.5} operating
point. Supplementary Fig.~\ref{fig:roc} shows the per-class ROC curves for one run and marks
that operating point, while Supplementary Table~\ref{tab:performance-code} provides the three-run summaries. Supplementary Table~\ref{tab:si-perclass}
compares the per-class AUROC of every model on both datasets, and
Extended Data Table~1 of the main text compares their per-class F1. The largest
external fixed-threshold degradation occurs for 1dAVb, discussed in
Supplementary Note~\ref{app:1davb-oppoint}.

\subsection{CODE-Test confusion counts}
\label{app:confusion}

Panel C of Supplementary Table~\ref{tab:performance-code} gives the true-positive,
false-positive, false-negative, and true-negative count ranges across the
three runs. The endpoints within a row can come from different seeds, so the
ranges do not define a single confusion matrix.

\section{Turning-point selection changes class trade-offs}
\label{app:selector-analysis}

Supplementary Table~\ref{tab:selector-perclass} reports the matched class means for
turning-point and uniform $k=30$ node selection. Turning-minus-uniform
macro-F1 differences are $+0.0028$, $+0.0202$, and $+0.0106$ in the three
paired seeds. The largest class-level F1 increase is for 1dAVb, where mean
recall rises from 0.583 to 0.702 even though AUPRC is lower. AF recall rises
from 0.949 to 1.000 and its AUPRC also rises. Uniform sampling has its largest
F1 advantage for SB. Both selectors attain recall 1.000 for SB, while uniform
sampling raises mean precision from 0.730 to 0.787. These results describe a
fixed-threshold trade-off rather than uniform improvement across metrics.

\section{All-points comparison}
\label{app:all-points}

All-points runs use the same seeds and effective batch size as the
turning-point runs. Because a physical batch of 128 all-points graphs exceeds
GPU memory, they use physical batches of eight with 16-step gradient
accumulation. Supplementary Table~\ref{tab:all-points-macro} reports the three-run means,
sample SDs, and paired within-seed differences.

At the fixed threshold, all points have lower macro F1 in every CODE-Test seed
and higher macro F1 in every PTB-XL seed. Their recall is lower in every paired
seed on both datasets. Higher PTB-XL precision accompanies the external F1
increase, while turning points have the in-domain F1 increase. The ranking
metric differences are smaller; among AUPRC and AUROC, only PTB-XL AUPRC has
the same paired sign in all three seeds. These three paired seeds support
descriptive replicated contrasts, not a significance or equivalence claim.

\section{Directed and bidirectional propagation}
\label{app:directionality}

Supplementary Table~\ref{tab:directionality-full} reports every ranking-metric contrast from
the matched directionality ablation. Differences are directed minus
bidirectional. The point estimates average the three within-seed differences,
and the paired bootstrap resamples shared exams while holding those seeds
fixed. The macro-F1 and AUPRC intervals include zero on both datasets, as does
the CODE-Test AUROC interval. PTB-XL AUROC is the one reported contrast whose
interval excludes zero, favoring directed propagation by
$0.000686\;[0.000172,0.001183]$.

\section{Computational efficiency}
\label{app:efficiency}

Supplementary Table~\ref{tab:computational-efficiency} reports graph construction, training,
and batched inference. We built the same first \num{1000}
PTB-XL exams with all 12 leads and 16 CPU workers;
node selection was the only changed factor. Across three builds on an AMD
Threadripper 7970X, all-points construction processed \num{39.0} exams/s and
turning-point construction processed \num{150.0} exams/s. Peak resident host
memory fell from \SI{875.4}{\mebi\byte} to \SI{275.4}{\mebi\byte}, and HDF5
size fell from \SI{576.7}{\mega\byte} to \SI{72.2}{\mega\byte}.

Runtime measurements use the full CODE100 turning-point HDF5, the reported
model dimensions, PyTorch 2.11, CUDA 13.0, BF16 autocast, and one RTX 5090.
Across 120 recorded epochs from three runs, training averaged
$132.0\pm3.7$ seconds per epoch. Complete runs, including validation, final
evaluation, and artifact logging, averaged $1.599\pm0.018$ hours. These are
measurements of the stated workstation and software stack.

\section{Delineated regional Shapley audit}
\label{app:regional-audit}

The structural audit uses a polarity-aware NeuroKit2 Prominence
delineator on the stored R anchors. Negative-polarity leads are inverted for
delineation. The selected model-output-blind configuration uses maximum QRS
interval \SI{204}{\milli\second}, maximum PR interval
\SI{500}{\milli\second}, maximum R-rise time
\SI{200}{\milli\second}, typical ST segment \SI{164}{\milli\second}, and
maximum P-, R-, and T-basepoint intervals of 100, 250, and
\SI{200}{\milli\second}, respectively. These marks provide technical
availability, not clinical fiducial ground truth.

For each lead-beat unit, PR runs from P onset to QRS onset, QRS from QRS onset
to QRS offset, and T from T onset to T offset. Region masks use the graph's
beat-relative timestamps. The shared PR/QRS boundary node belongs to QRS so
that the masks remain disjoint. Deleting a region removes its nodes and their
incident edges and then links the last remaining node before the region to
the first remaining node after it, so that the beat graph stays connected. The
new edge carries the time and amplitude differences between its two nodes,
as every other edge does. Units outside the active availability rule remain
intact.

The three regions define all $2^3=8$ structural states. For each exam and
trained seed, one state evaluation supplies the outputs for 1dAVb, RBBB, and
LBBB. Exact Shapley allocation decomposes the intact-minus-all-regions-deleted
logit difference among PR, QRS, and T. The region with the largest logit
contribution is the verdict leader. Correct classification is defined from the
intact probability at the fixed \num{0.5} threshold. Thus, statements about
expected-region leadership among correct verdicts condition on that threshold
and do not cover misclassified verdicts.

The main-text fixed-window comparison changes only the region boundaries
on the same available beats. Both fixed protocols use PR from
\(-0.35\) to \(-0.07\)\,s and QRS from \(-0.06\) to \(+0.12\)\,s relative
to the stored R anchor. The class-specific T window runs from \(+0.15\) to
\(+0.43\)\,s for 1dAVb and from \(+0.15\) to \(+0.33\)\,s for RBBB and
LBBB. The standardized T window runs from \(+0.15\) to \(+0.45\)\,s for
all three classes. Each mask selects nodes only within its own beat.

\section{Delineation availability sensitivity}
\label{app:availability}

The primary beat-complete rule includes a lead-beat unit only when PR, QRS,
and T are all marked and ordered. Its three region masks therefore use the
same units. The per-region sensitivity rule includes each marked region
independently. A unit missing only T can contribute PR and QRS under the
per-region rule, while its T nodes remain intact in every state. The rules
change the deletion support, so their results are separate sensitivity
analyses rather than ordered limits on an underlying effect.

Panel A of Supplementary Table~\ref{tab:si-cells} reports both availability rules for the full
label-positive cohort and the morphology-isolated subset. In the full cohort,
LBBB expected-region leadership among correct verdicts changes from 78/83 to
82/83, 1dAVb changes from 47/59 to 45/59, and RBBB changes from 95/96 to 96/96. These are
measured sensitivity differences. They do not identify the cause of an
individual changed verdict. The morphology-isolated subset excludes concurrent
positives among 1dAVb, RBBB, and LBBB only. It does not exclude the other task
labels or unmeasured conditions, so it limits one source of co-labeling rather
than defining single-disease exams.

Panel B of Supplementary Table~\ref{tab:si-cells} applies minimum beat-complete coverage filters and
then recomputes strict all-seed agreement. For LBBB, the strict counts are
26/30 with no floor, 25/29 at 0.5, 25/27 at 0.6, 24/26 at 0.7, and 20/22 at
0.8. Both numerator and denominator change with the retained cohort, and the
agreement fraction is not monotonic across these rows. The floor analysis
therefore measures cohort sensitivity; it does not establish causal dilution
by missing delineations.

\begin{figure}[p]
  \centering
  \includegraphics[width=\linewidth,height=0.78\textheight,keepaspectratio]{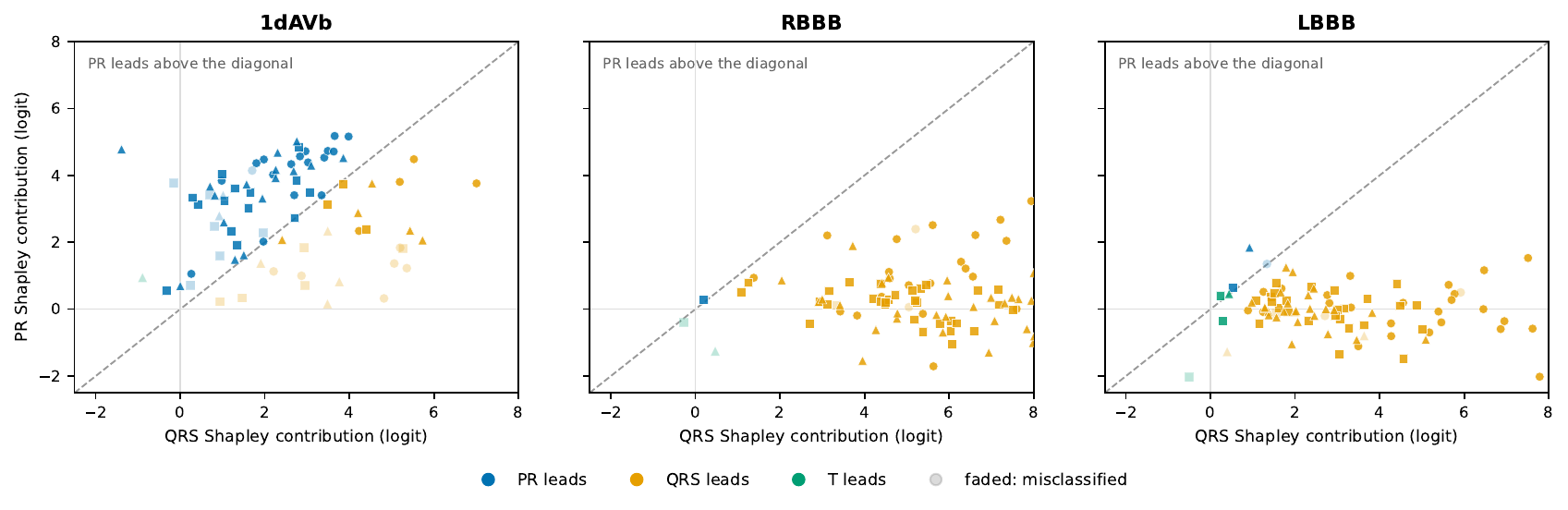}
  \caption{\textbf{PR and QRS Shapley contributions per verdict.} PR against QRS Shapley contribution for every verdict, primary
  availability rule (\(N=276\) verdicts: 84 for 1dAVb, 102 for RBBB, and
  90 for LBBB). Color marks the leading region; faded points are
  misclassified verdicts; markers are the three seeds. The dashed line is
  the PR-equals-QRS diagonal. Axes cross zero because Shapley contributions
  are signed.}
    \label{fig:si-scatter}
\end{figure}

Supplementary Fig.~\ref{fig:si-scatter} plots PR against QRS Shapley contributions for all
276 label-positive verdicts and fades the misclassified points. The reported
leadership rates instead condition on the 238 correctly classified verdicts.
Within that correct-classification subset, PR leads 47/59 1dAVb verdicts, QRS
leads 95/96 RBBB verdicts, and QRS leads 78/83 LBBB verdicts under the primary
rule. The 1dAVb points near the PR-equals-QRS diagonal have similar
contributions from the two regions. This plot describes model sensitivity
under the stated intervention, not physiological causality.

\section{Paired baseline comparisons}
\label{app:baseline-bootstrap}

Supplementary Table~\ref{tab:baseline-bootstrap} gives the paired macro-F1 comparisons with the baseline reruns.

\section{External 1dAVb fixed-threshold limitation}
\label{app:1davb-oppoint}

Across the three strict-SCP PTB-XL evaluations in
panel B of Supplementary Table~\ref{tab:performance-code}, 1dAVb F1 at the fixed \num{0.5} threshold is
$0.440878\pm0.017711$. Precision ranges from \num{0.301} to \num{0.350}, while
recall ranges from \num{0.671} to \num{0.736}. AUROC remains
$0.943680\pm0.002832$. The three-seed result therefore shows a gap between
ranking and the transported fixed operating point under the declared strict
SCP-code mapping. It does not determine whether the gap arises from score
calibration, cohort shift, or differences between the SCP-code and CODE label
protocols.

\section{External SB label protocol}
\label{app:sb-protocol}

CODE labels SB at a heart rate of 50~beats/min or less, while the PTB-XL
\texttt{SBRAD} statement covers sinus rhythm below 60~beats/min.
Supplementary Table~\ref{tab:sb-protocol} stratifies strict-SCP PTB-XL by the heart rate
measured from the stored R anchors. SB recall is 0.97 below 50~beats/min and
falls to 0.07 between 55 and 60~beats/min, where the CODE protocol would not
assign SB. Below 50~beats/min, the model also calls SB in 73\% of the 105 exams
without \texttt{SBRAD}. The external SB recall of 0.667 therefore reflects the
label definition more than a detection failure. ST shows the same agreement with
its threshold: recall is 0.92 at 100--105~beats/min, and exams without
\texttt{STACH} are called ST below 100~beats/min in only 0.2\% of verdicts.

\section{Timing interventions}
\label{app:timing}

\begin{figure}[p]
  \centering
  \includegraphics[width=\linewidth,height=0.78\textheight,keepaspectratio]{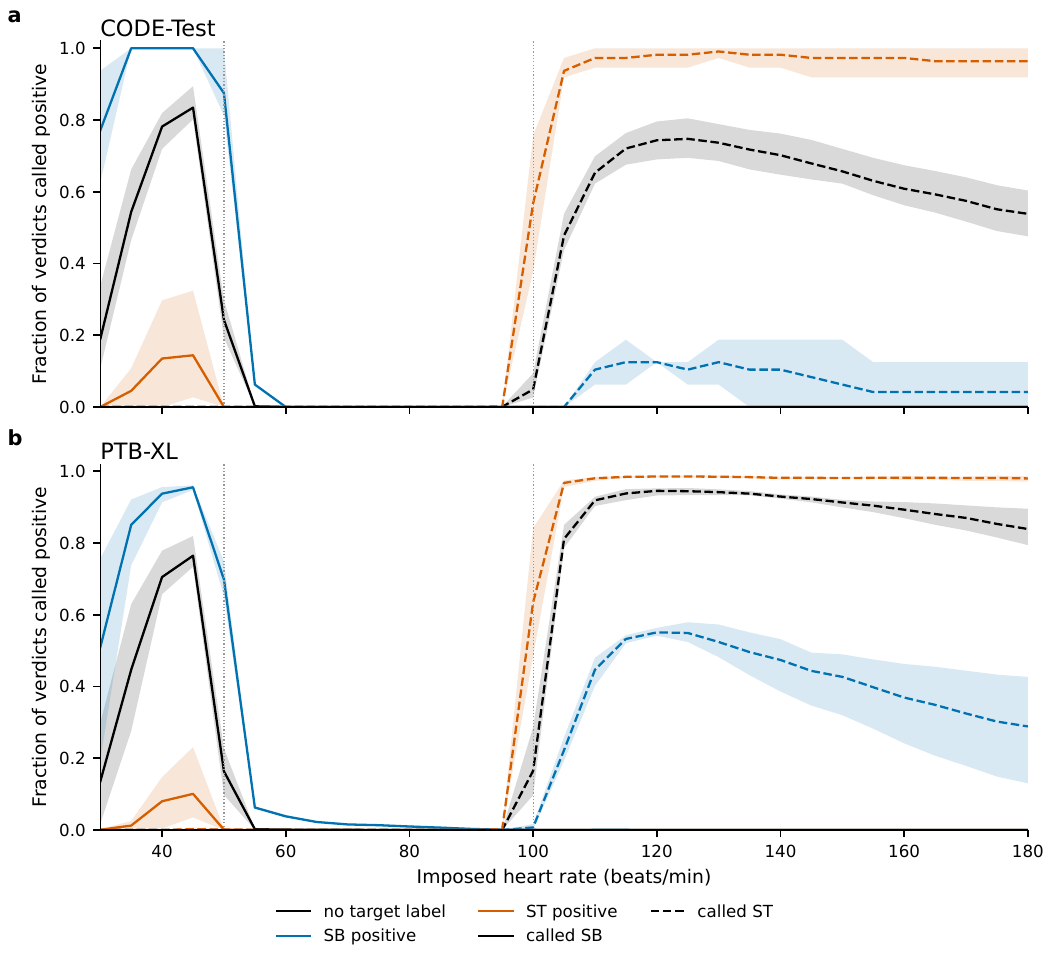}
  \caption{\textbf{Uniform rate sweep on both datasets.} Fraction of verdicts called
  SB (solid) or ST (dashed) as every beat's time axis is scaled to the imposed
  heart rate. Lines are means over three seeds, and bands span the seeds.}
    \label{fig:si-rate-sweep}
\end{figure}

\begin{figure}[p]
  \centering
  \includegraphics[width=\linewidth,height=0.78\textheight,keepaspectratio]{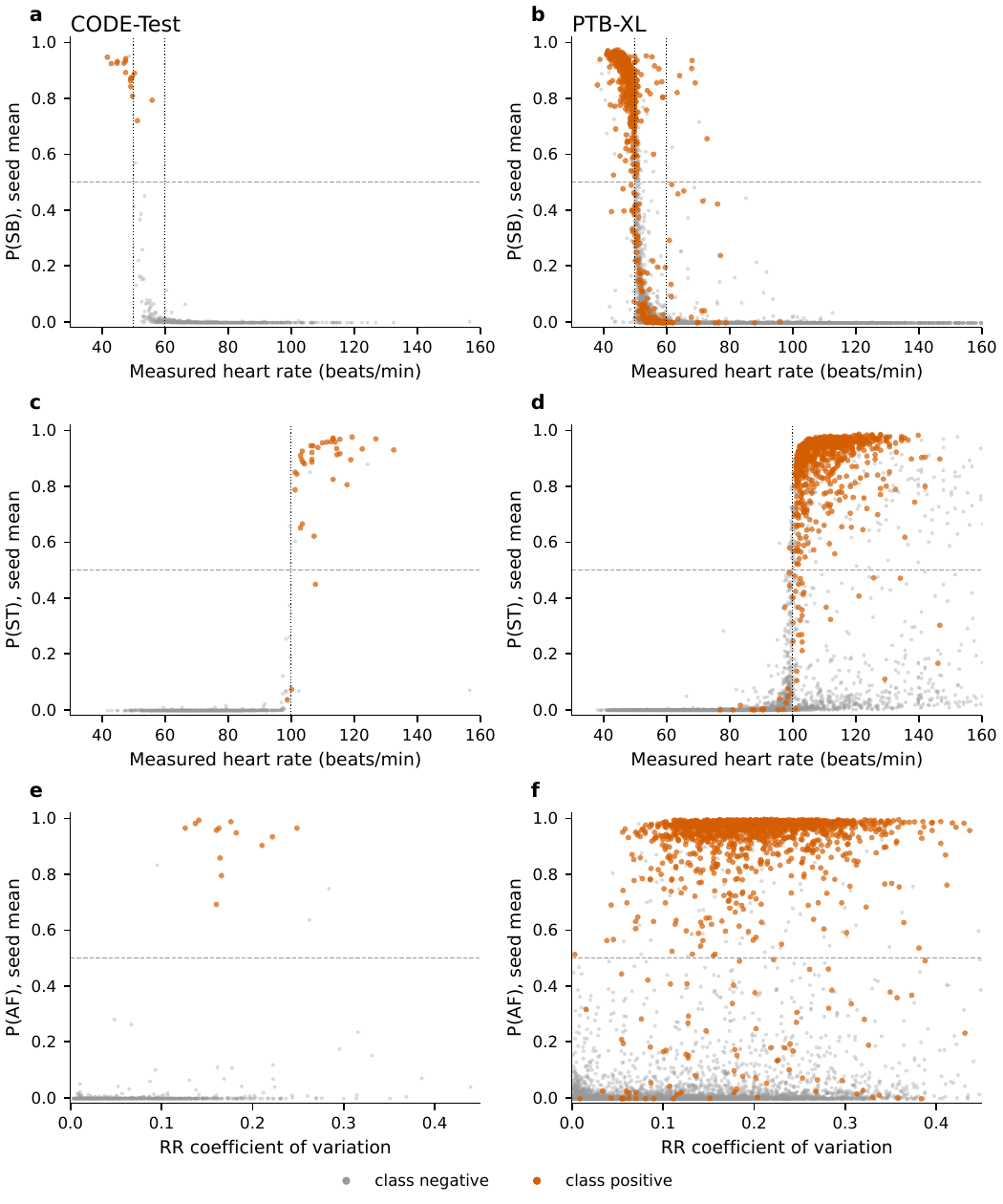}
  \caption{\textbf{Intact predictions against measured rhythm descriptors.} Seed-mean
  probability of SB and ST against the heart rate measured from the stored R
  anchors, and of AF against the RR coefficient of variation. Dotted lines mark
  50 and 60~beats/min for SB and 100~beats/min for ST.}
    \label{fig:si-rate-descriptors}
\end{figure}

Supplementary Table~\ref{tab:si-timing} adds 95\% exam-bootstrap intervals to the sequence
interventions of the main text and reports the template with the original beat
durations restored. Supplementary Table~\ref{tab:si-rate} compares the uniform and the
interval-preserving rate change. Supplementary Fig.~\ref{fig:si-rate-sweep} shows the
uniform sweep on both datasets, and Supplementary Fig.~\ref{fig:si-rate-descriptors} plots
the intact predictions against the measured heart rate and RR variation.
Under uniform scaling, AF-positive PTB-XL verdicts stay AF in 0.81
[0.80, 0.82] of cases at an imposed 40~beats/min and 0.90 [0.89, 0.91] at
180~beats/min, so the AF call does not follow rate. Uniform stretching to
40~beats/min also turns 56\% of CODE-Test exams without any target label into
1dAVb calls, which is consistent with a model that reads the absolute PR
duration.

\section{Lead Shapley values}
\label{app:leads}

\begin{figure}[p]
  \centering
  \includegraphics[width=\linewidth,height=0.78\textheight,keepaspectratio]{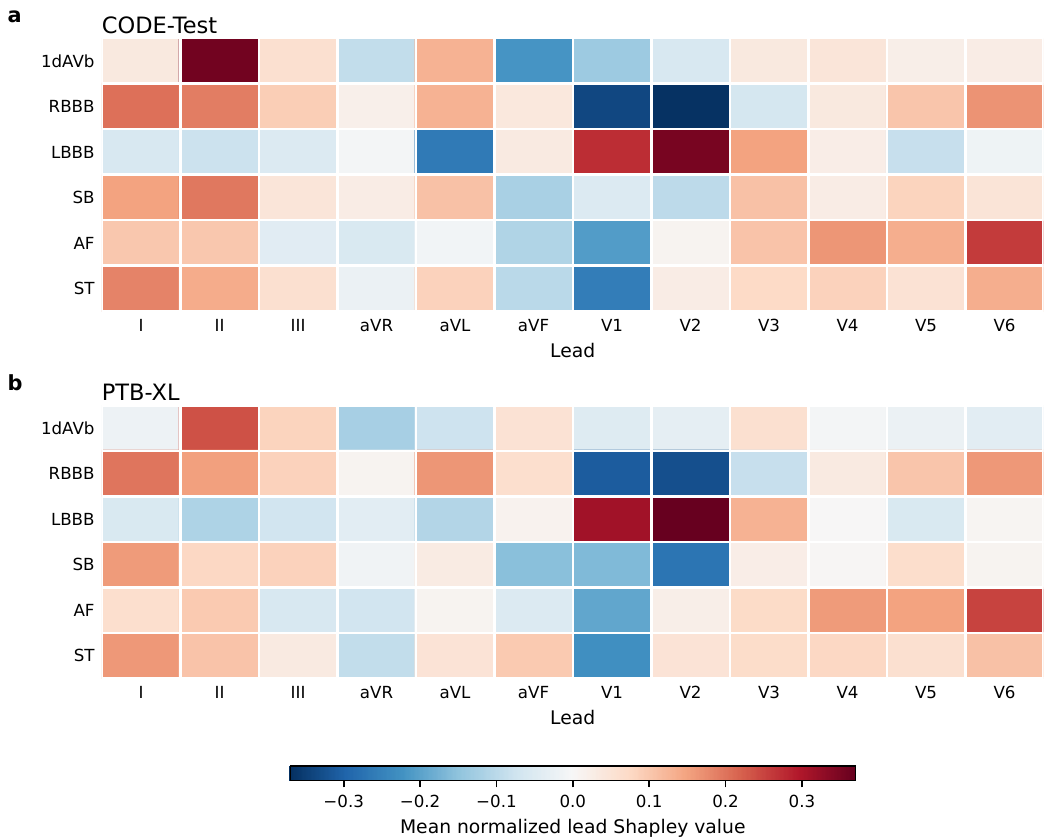}
  \caption{\textbf{Mean normalized lead Shapley value per label.} Each cell averages,
  over label-positive verdicts, a lead's Shapley value divided by the sum of
  positive lead values. Negative cells can reflect dilution under attention
  pooling.}
    \label{fig:si-leads}
\end{figure}

Supplementary Fig.~\ref{fig:si-leads} shows the mean normalized Shapley value of each lead
for each label, and Supplementary Table~\ref{tab:si-leads} reports how often one lead alone
keeps a correct call. The six limb leads are linear combinations of two
independent signals. The exam embedding is an attention-weighted mean of lead
embeddings, so a negative Shapley value can reflect dilution of stronger leads
rather than evidence against the label.

\section{Regional deletion for the rhythm labels}
\label{app:rhythm-regions}

Supplementary Table~\ref{tab:si-rhythm-regions} applies the eight-state PR--QRS--T game of the
main text to SB, AF, and ST. These labels have no expected waveform region, and
deleting a region does not remove rate information, because \(t_{\mathrm{norm}}\)
and the remaining \(\Delta t\) values still encode beat duration. The table
therefore shows where the morphological evidence sits, while the timing interventions
test the rhythm calls. The delineator marks a PR region in 96.5\% of AF lead--beat units, so for
AF this region is the pre-QRS interval rather than a P wave.

\clearpage
\clearpage

\clearpage

\clearpage
\begin{table}
  \renewcommand{\arraystretch}{1.4}
  \centering
  \caption{\textbf{Data splits after graph quality control, with exam counts, beat-graph
  counts, and per-class positives.} CODE100 training and validation form a
  disjoint \num{98}/\num{2}\% patient-hash split; validation only selects
  checkpoints. CODE-Test and PTB-XL are used only for evaluation. Multilabel
  class counts may exceed the exam count. The last row gives the strict PTB-XL
  SCP-ECG code mapped to each class by code presence: first-degree block only,
  complete bundle-branch blocks, and AF without atrial flutter.}
  \label{tab:prevalence}
  \fontsize{8}{10}\selectfont
  \begin{adjustbox}{max width=\linewidth,center}
  \begin{tabular}{llrrrrrrrr}
    \toprule
    Dataset & Role & Exams & Beat graphs & 1dAVb & RBBB & LBBB & SB & AF & ST \\
    \midrule
    CODE100 training   & Training             & \num{2237125} & \num{279394549} & \num{34587} & \num{61558} & \num{35843} & \num{36642} & \num{40357} & \num{48185} \\
    CODE100 validation & Checkpoint selection & \num{46378}   & \num{5818784}   & \num{714}   & \num{1268}  & \num{731}   & \num{749}   & \num{829}   & \num{985} \\
    CODE-Test          & In-domain test       & \num{827}     & \num{106340}    & \num{28}    & \num{34}    & \num{30}    & \num{16}    & \num{13}    & \num{37} \\
    PTB-XL             & External test        & \num{21799}   & \num{3176537}   & \num{793}   & \num{541}   & \num{536}   & \num{637}   & \num{1514}  & \num{826} \\
    \midrule
    PTB-XL SCP code    & & & & \texttt{1AVB} & \texttt{CRBBB} & \texttt{CLBBB} & \texttt{SBRAD} & \texttt{AFIB} & \texttt{STACH} \\
    \bottomrule
  \end{tabular}
  \end{adjustbox}
\end{table}

\begin{table}
  \renewcommand{\arraystretch}{1.4}
  \caption{\textbf{Paired macro-F1 comparisons with baselines.}
  Values are the macro F1 of LYNX-ECG minus the comparator's, followed by a
  95\% percentile confidence interval from \num{2000} shared-exam
  bootstrap resamples. The trained seed pairs are held fixed.}
  \label{tab:baseline-bootstrap}
  \centering
  \fontsize{8}{10}\selectfont
  \begin{adjustbox}{max width=\linewidth,center}
  \begin{tabular}{lcc}
    \toprule
    Comparator & CODE-Test & PTB-XL strict-SCP \\
    \midrule
    DNN~\cite{ribeiro2020} & \(+0.0399\;[+0.0136,+0.0704]\) & \(+0.0117\;[+0.0065,+0.0172]\) \\
    ViT~\cite{dosovitskiy2021vit} & \(+0.0245\;[-0.0040,+0.0562]\) & \(+0.0150\;[+0.0100,+0.0197]\) \\
    EfficientNet~\cite{tan2019efficientnet} & \(+0.0168\;[-0.0136,+0.0435]\) & \(+0.0025\;[-0.0023,+0.0074]\) \\
    ECGFounder-Scratch~\cite{li2024ecgfounder} & \(+0.0256\;[+0.0009,+0.0509]\) & \(-0.0004\;[-0.0042,+0.0036]\) \\
    ECGFounder-Finetuned~\cite{li2024ecgfounder} & \(+0.0315\;[+0.0010,+0.0632]\) & \(-0.0044\;[-0.0090,+0.0002]\) \\
    \bottomrule
  \end{tabular}
  \end{adjustbox}
\end{table}

\begin{table}
  \renewcommand{\arraystretch}{1.4}
  \centering
  \caption{\textbf{SB and ST recall and false-call rate on strict-SCP PTB-XL by the
  measured heart rate.} Values pool three seeds; brackets give 95\%
  exam-bootstrap intervals.}
  \label{tab:sb-protocol}
  \fontsize{8}{10}\selectfont
  \begin{adjustbox}{max width=\linewidth,center}
  \begin{tabular}{llrcrc}
    \toprule
    Label & Heart rate (beats/min) & Positives & Recall & Negatives & False-call rate \\
    \midrule
    SB & $<50$     & 373 & 0.97 [0.95, 0.98] & 105   & 0.73 [0.66, 0.81] \\
       & 50--55    & 123 & 0.40 [0.31, 0.48] & 1074  & 0.07 [0.06, 0.08] \\
       & 55--60    & 108 & 0.07 [0.03, 0.13] & 2011  & 0.00 [0.00, 0.00] \\
       & $\geq 60$ & 33  & 0.24 [0.12, 0.37] & 17972 & 0.00 [0.00, 0.00] \\
    \midrule
    ST & $<100$    & 20  & 0.10 [0.02, 0.20] & 20125 & 0.00 [0.00, 0.00] \\
       & 100--105  & 273 & 0.92 [0.89, 0.95] & 243   & 0.40 [0.34, 0.45] \\
       & 105--110  & 189 & 1.00 [0.99, 1.00] & 105   & 0.23 [0.16, 0.31] \\
       & $\geq 110$& 344 & 0.98 [0.97, 0.99] & 500   & 0.30 [0.26, 0.34] \\
    \bottomrule
  \end{tabular}
  \end{adjustbox}
\end{table}

\begin{table}
  \renewcommand{\arraystretch}{1.4}
  \centering
  \caption{\textbf{Performance of LYNX-ECG across three seeds at the fixed 0.5 threshold
  on CODE-Test (panel A) and strict-SCP PTB-XL (panel B).} Per-class AUROC, F1,
  precision, and recall are ranges, whereas per-class AUPRC and all macro
  metrics are means $\pm$ sample SD. Panel C gives the CODE-Test per-class
  confusion-count ranges. Range endpoints need not come from the same seed.}
  \label{tab:performance-code}
  \fontsize{8}{10}\selectfont
  \textbf{A. CODE-Test}\par\smallskip
  \begin{adjustbox}{max width=\linewidth,center}
  \begin{tabular}{lccccc}
    \toprule
    Class & AUROC & AUPRC & F1 & Precision & Recall \\
    \midrule
    1dAVb & 0.9953--0.9964 & $0.8945\pm0.0019$ & 0.696--0.830 & 0.880--0.913 & 0.571--0.786 \\
    RBBB  & 0.9987--0.9991 & $0.9785\pm0.0034$ & 0.877--0.941 & 0.821--0.941 & 0.941 \\
    LBBB  & 0.9998--1.0000 & $0.9980\pm0.0027$ & 0.947--0.966 & 1.000 & 0.900--0.933 \\
    SB    & 0.9978--0.9983 & $0.8390\pm0.0544$ & 0.800--0.865 & 0.667--0.762 & 1.000 \\
    AF    & 0.9994--0.9999 & $0.9777\pm0.0149$ & 0.897--0.929 & 0.812--0.867 & 1.000 \\
    ST    & 0.9979--0.9984 & $0.9638\pm0.0033$ & 0.895--0.921 & 0.872--0.943 & 0.892--0.946 \\
    \midrule
    macro & $0.998415\pm0.000050$ & $0.9419\pm0.0124$ & $0.8890\pm0.0035$ & $0.8796\pm0.0158$ & $0.9141\pm0.0156$ \\
    \bottomrule
  \end{tabular}
  \end{adjustbox}

  \medskip
  \textbf{B. Strict-SCP PTB-XL}\par\smallskip
  \begin{adjustbox}{max width=\linewidth,center}
  \begin{tabular}{lccccc}
    \toprule
    Class & AUROC & AUPRC & F1 & Precision & Recall \\
    \midrule
    1dAVb & 0.9405--0.9458 & $0.4034\pm0.0102$ & 0.425--0.460 & 0.301--0.350 & 0.671--0.736 \\
    RBBB  & 0.9948--0.9955 & $0.7713\pm0.0130$ & 0.740--0.793 & 0.593--0.677 & 0.957--0.985 \\
    LBBB  & 0.9966--0.9971 & $0.9361\pm0.0024$ & 0.706--0.774 & 0.551--0.644 & 0.970--0.979 \\
    SB    & 0.9623--0.9651 & $0.6957\pm0.0030$ & 0.691--0.708 & 0.713--0.764 & 0.659--0.670 \\
    AF    & 0.9940--0.9944 & $0.9446\pm0.0023$ & 0.919--0.923 & 0.891--0.919 & 0.927--0.948 \\
    ST    & 0.9949--0.9951 & $0.8708\pm0.0042$ & 0.791--0.829 & 0.676--0.747 & 0.931--0.953 \\
    \midrule
    macro & $0.981478\pm0.000524$ & $0.7703\pm0.0043$ & $0.7300\pm0.0012$ & $0.6513\pm0.0014$ & $0.8673\pm0.0054$ \\
    \bottomrule
  \end{tabular}
  \end{adjustbox}

  \medskip
  \textbf{C. CODE-Test confusion counts}\par\smallskip
  \begin{adjustbox}{max width=\linewidth,center}
  \begin{tabular}{lrrrr}
    \toprule
    Class & TP & FP & FN & TN \\
    \midrule
    1dAVb & 16--22 & 2--3 & 6--12 & 796--797 \\
    RBBB  & 32     & 2--7 & 2     & 786--791 \\
    LBBB  & 27--28 & 0    & 2--3  & 797 \\
    SB    & 16     & 5--8 & 0     & 803--806 \\
    AF    & 13     & 2--3 & 0     & 811--812 \\
    ST    & 33--35 & 2--5 & 2--4  & 785--788 \\
    \bottomrule
  \end{tabular}
  \end{adjustbox}
\end{table}

\begin{table}
  \renewcommand{\arraystretch}{1.3}
  \centering
  \caption{\textbf{Per-class AUROC of every model on CODE-Test and strict-SCP PTB-XL.}
  Values are means $\pm$ sample SD across three training runs. Boldface marks
  the highest mean and underline marks the second-highest mean in each column
  and dataset. The macro column repeats Table~1 of the main text.
  Extended Data Table~1 of the main text gives the per-class F1.}
  \label{tab:si-perclass}
  \fontsize{8}{10}\selectfont
  \begin{adjustbox}{max width=\linewidth,center}
  \begin{tabular}{@{}lccccccc@{}}
    \toprule
    Model & 1dAVb & RBBB & LBBB & SB & AF & ST & Macro \\
    \midrule
    \multicolumn{8}{@{}l}{\textit{CODE-Test ($n=827$)}} \\
    LYNX-ECG & $\underline{0.9959\pm0.0006}$ & $0.9989\pm0.0002$ & $\mathbf{0.9999\pm0.0001}$ & $\underline{0.9980\pm0.0003}$ & $0.9996\pm0.0003$ & $\mathbf{0.9982\pm0.0002}$ & $\mathbf{0.9984\pm0.0001}$ \\
    DNN~\cite{ribeiro2020} & $0.9952\pm0.0018$ & $0.9989\pm0.0003$ & $0.9990\pm0.0006$ & $0.9973\pm0.0002$ & $0.9985\pm0.0008$ & $0.9968\pm0.0001$ & $0.9976\pm0.0002$ \\
    ViT~\cite{dosovitskiy2021vit} & $0.9913\pm0.0035$ & $\underline{0.9990\pm0.0001}$ & $\mathbf{0.9999\pm0.0001}$ & $0.9974\pm0.0003$ & $\mathbf{0.9999\pm0.0001}$ & $0.9950\pm0.0003$ & $0.9971\pm0.0006$ \\
    EfficientNet~\cite{tan2019efficientnet} & $\mathbf{0.9961\pm0.0010}$ & $\mathbf{0.9991\pm0.0003}$ & $\mathbf{0.9999\pm0.0001}$ & $0.9970\pm0.0008$ & $\underline{0.9998\pm0.0003}$ & $0.9953\pm0.0005$ & $\underline{0.9979\pm0.0003}$ \\
    ECGFounder-Scratch~\cite{li2024ecgfounder} & $0.9934\pm0.0007$ & $0.9985\pm0.0006$ & $\underline{0.9998\pm0.0002}$ & $\mathbf{0.9985\pm0.0004}$ & $0.9997\pm0.0003$ & $\underline{0.9978\pm0.0006}$ & $\underline{0.9979\pm0.0003}$ \\
    ECGFounder-Finetuned~\cite{li2024ecgfounder} & $0.9875\pm0.0083$ & $0.9983\pm0.0011$ & $0.9995\pm0.0003$ & $0.9955\pm0.0014$ & $0.9986\pm0.0011$ & $0.9972\pm0.0010$ & $0.9961\pm0.0016$ \\
    \midrule
    \multicolumn{8}{@{}l}{\textit{PTB-XL ($n=21{,}799$)}} \\
    LYNX-ECG & $0.9437\pm0.0028$ & $\underline{0.9951\pm0.0003}$ & $\mathbf{0.9968\pm0.0002}$ & $\underline{0.9640\pm0.0015}$ & $\mathbf{0.9942\pm0.0002}$ & $\mathbf{0.9950\pm0.0001}$ & $\underline{0.9815\pm0.0005}$ \\
    DNN & $\underline{0.9472\pm0.0037}$ & $\mathbf{0.9957\pm0.0002}$ & $\underline{0.9956\pm0.0012}$ & $0.9631\pm0.0033$ & $\underline{0.9939\pm0.0005}$ & $0.9901\pm0.0005$ & $0.9809\pm0.0008$ \\
    ViT & $0.9348\pm0.0070$ & $0.9941\pm0.0005$ & $0.9945\pm0.0004$ & $\mathbf{0.9666\pm0.0015}$ & $0.9921\pm0.0006$ & $0.9924\pm0.0022$ & $0.9791\pm0.0018$ \\
    EfficientNet & $0.9364\pm0.0062$ & $0.9950\pm0.0004$ & $0.9954\pm0.0009$ & $0.9622\pm0.0017$ & $0.9932\pm0.0007$ & $0.9916\pm0.0004$ & $0.9790\pm0.0013$ \\
    ECGFounder-Scratch & $\mathbf{0.9473\pm0.0071}$ & $0.9945\pm0.0006$ & $\underline{0.9956\pm0.0009}$ & $\mathbf{0.9666\pm0.0035}$ & $0.9937\pm0.0002$ & $\underline{0.9930\pm0.0016}$ & $\mathbf{0.9818\pm0.0007}$ \\
    ECGFounder-Finetuned & $0.9395\pm0.0048$ & $0.9937\pm0.0003$ & $0.9932\pm0.0018$ & $0.9281\pm0.0025$ & $0.9915\pm0.0001$ & $0.9921\pm0.0004$ & $0.9730\pm0.0010$ \\
    \bottomrule
  \end{tabular}
  \end{adjustbox}
\end{table}

\begin{table}
  \renewcommand{\arraystretch}{1.4}
  \centering
  \caption{\textbf{Matched CODE-Test class means across seeds.}
  $\Delta$ is turning-point selection minus uniform $k=30$; positive values
  favor turning-point selection.}
  \label{tab:selector-perclass}
  \fontsize{8}{10}\selectfont
  \begin{adjustbox}{max width=\linewidth,center}
  \begin{tabular}{lrrrrrr}
    \toprule
    & \multicolumn{3}{c}{F1@0.5} & \multicolumn{3}{c}{AUPRC} \\
    \cmidrule(lr){2-4}\cmidrule(l){5-7}
    Class & Turning & Uniform & $\Delta$ & Turning & Uniform & $\Delta$ \\
    \midrule
    1dAVb & 0.783 & 0.715 & $\mathbf{+0.068}$ & 0.895 & 0.915 & $-0.020$ \\
    RBBB  & 0.920 & 0.937 & $-0.017$ & 0.978 & 0.979 & $-0.001$ \\
    LBBB  & 0.959 & 0.954 & $+0.006$ & 0.998 & 0.999 & $-0.001$ \\
    SB    & 0.843 & 0.881 & $-0.038$ & 0.839 & 0.920 & $-0.081$ \\
    AF    & 0.918 & 0.871 & $\mathbf{+0.047}$ & 0.978 & 0.960 & $\mathbf{+0.018}$ \\
    ST    & 0.911 & 0.911 & $+0.000$ & 0.964 & 0.963 & $+0.001$ \\
    \bottomrule
  \end{tabular}
  \end{adjustbox}
\end{table}

\begin{table}
  \renewcommand{\arraystretch}{1.4}
  \centering
  \caption{\textbf{Three-seed all-points comparison as mean $\pm$ sample SD.} Each
  paired difference uses values from the same seed and is all points minus
  turning points. For F1, this difference was negative on CODE-Test and
  positive on PTB-XL in each of the three seeds.}
  \label{tab:all-points-macro}
  \fontsize{8}{10}\selectfont
  \begin{adjustbox}{max width=\linewidth,center}
  \begin{tabular}{llrrr}
    \toprule
    Dataset & Metric & All points & Turning points & Paired $\Delta$ \\
    \midrule
    CODE-Test
      & F1 & $0.869977\pm0.010829$ & $0.889040\pm0.003517$ & $-0.019063\pm0.007336$ \\
      & AUPRC & $0.950773\pm0.007045$ & $0.941906\pm0.012364$ & $+0.008866\pm0.009120$ \\
      & AUROC & $0.998528\pm0.000171$ & $0.998415\pm0.000050$ & $+0.000113\pm0.000164$ \\
      & Precision & $0.880707\pm0.020565$ & $0.879622\pm0.015825$ & $+0.001085\pm0.006775$ \\
      & Recall & $0.884078\pm0.025188$ & $0.914116\pm0.015607$ & $-0.030039\pm0.014806$ \\
      & ECE & $0.007952\pm0.000709$ & $0.007992\pm0.001015$ & $-0.000041\pm0.001655$ \\
    \midrule
    PTB-XL
      & F1 & $0.742880\pm0.005454$ & $0.730007\pm0.001231$ & $+0.012873\pm0.006673$ \\
      & AUPRC & $0.775312\pm0.004909$ & $0.770333\pm0.004325$ & $+0.004980\pm0.001035$ \\
      & AUROC & $0.981766\pm0.000774$ & $0.981478\pm0.000524$ & $+0.000288\pm0.000829$ \\
      & Precision & $0.672100\pm0.009546$ & $0.651333\pm0.001351$ & $+0.020768\pm0.008216$ \\
      & Recall & $0.855815\pm0.005375$ & $0.867254\pm0.005354$ & $-0.011439\pm0.001345$ \\
      & ECE & $0.014368\pm0.001549$ & $0.017375\pm0.001179$ & $-0.003007\pm0.000374$ \\
    \bottomrule
  \end{tabular}
  \end{adjustbox}
\end{table}

\begin{table}
  \renewcommand{\arraystretch}{1.4}
  \centering
  \caption{\textbf{Measured computational efficiency.} Each graph build used the same
  1,000 exams, three builds, and 16 CPU workers; memory is peak RSS and output
  size is the resulting HDF5 file. Reported training values cover three seeds,
  40 epochs, and 454,473 exams per epoch; full-run time includes validation,
  final evaluation, and artifact logging. Batched inference used 40 batches of
  128 exams on one RTX 5090. ``Allocated'' is peak framework-allocated GPU
  memory. The loader used page-cached data on local scratch.}
  \label{tab:computational-efficiency}
  \fontsize{8}{10}\selectfont
  \begin{adjustbox}{max width=\linewidth,center}
  \begin{tabular}{llll}
    \toprule
    Stage & Time or throughput & Output size & Peak memory \\
    \midrule
    Build, all points
      & $25.663\pm0.095$ s; \num{39.0} exams/s
      & 576.7 MB HDF5
      & \SI{875.4}{\mebi\byte} RSS \\
    Build, turning points
      & $6.667\pm0.006$ s; \num{150.0} exams/s
      & 72.2 MB HDF5
      & \SI{275.4}{\mebi\byte} RSS \\
    Reported training epochs
      & $132.0\pm3.7$ s/epoch; \num{3444} exams/s
      & N/A
      & N/A \\
    Reported full runs
      & $1.599\pm0.018$ h/run
      & N/A
      & N/A \\
    Batched inference
      & 31.9 ms/batch; \num{4016} exams/s
      & N/A
      & \SI{3.36}{\gibi\byte} allocated \\
    \bottomrule
  \end{tabular}
  \end{adjustbox}
\end{table}

\begin{table}
  \renewcommand{\arraystretch}{1.4}
  \centering
  \caption{\textbf{Directed and bidirectional ranking metrics.} $\Delta$ is directed
  minus bidirectional; brackets give 95\% paired exam-bootstrap intervals.}
  \label{tab:directionality-full}
  \fontsize{8}{10}\selectfont
  \begin{adjustbox}{max width=\linewidth,center}
  \begin{tabular}{llrrr}
    \toprule
    Dataset & Metric & Directed & Bidirectional & $\Delta$ [95\% interval] \\
    \midrule
    CODE-Test
      & F1 & 0.889040 & 0.875832 & $+0.013208\;[-0.006583,+0.032856]$ \\
      & AUROC & 0.998415 & 0.998441 & $-0.000026\;[-0.000540,+0.000389]$ \\
      & AUPRC & 0.941906 & 0.950522 & $-0.008616\;[-0.020736,+0.007055]$ \\
    \midrule
    PTB-XL
      & F1 & 0.730007 & 0.732977 & $-0.002970\;[-0.006064,+0.000208]$ \\
      & AUROC & 0.981478 & 0.980793 & $+0.000686\;[+0.000172,+0.001183]$ \\
      & AUPRC & 0.770333 & 0.772413 & $-0.002080\;[-0.006112,+0.001808]$ \\
    \bottomrule
  \end{tabular}
  \end{adjustbox}
\end{table}

\begin{table}
  \renewcommand{\arraystretch}{1.4}
  \centering
  \caption{\textbf{PR--QRS--T structural deletion for the rhythm labels under the
  beat-complete rule, pooled over three seeds.} Counts give the leading region
  among correct verdicts; Shapley values are medians [IQR] in logits over all
  positive verdicts. ``Complete'' is the fraction of lead--beat units with all
  three regions delineated. These labels have no expected region, so the
  leading region does not explain the rhythm call.}
  \label{tab:si-rhythm-regions}
  \fontsize{8}{10}\selectfont
  \begin{adjustbox}{max width=\linewidth,center}
  \begin{tabular}{llcccccccc}
    \toprule
    Dataset & Label & Complete & Correct & PR leads & QRS leads & T leads & PR & QRS & T \\
    \midrule
    CODE-Test & SB & 0.92 & 48/48    & 1    & 45   & 2  & 0.55 [0.28, 1.28] & 3.83 [2.84, 4.52] & 1.60 [1.22, 2.01] \\
              & AF & 0.80 & 39/39    & 0    & 39   & 0  & 1.08 [0.58, 1.94] & 4.14 [2.77, 5.32] & $-0.65$ [$-1.08$, 0.02] \\
              & ST & 0.73 & 101/111  & 23   & 78   & 0  & 0.40 [0.22, 0.59] & 1.04 [0.37, 1.85] & $-0.72$ [$-1.12$, $-0.31$] \\
    \midrule
    PTB-XL    & SB & 0.92 & 1274/1911 & 27   & 1152 & 95 & 0.84 [0.10, 1.47] & 2.63 [0.81, 4.27] & 1.55 [1.06, 2.29] \\
              & AF & 0.65 & 4258/4542 & 472  & 3747 & 39 & 0.90 [0.34, 1.54] & 2.76 [1.40, 4.30] & $-0.45$ [$-0.90$, $-0.08$] \\
              & ST & 0.63 & 2335/2478 & 595  & 1719 & 21 & 0.24 [0.05, 0.49] & 0.62 [0.16, 1.26] & $-0.38$ [$-0.80$, $-0.14$] \\
    \bottomrule
  \end{tabular}
  \end{adjustbox}
\end{table}

\begin{table}
  \renewcommand{\arraystretch}{1.4}
  \centering
  \caption{\textbf{Sensitivity of expected-region verdicts to the availability rule,
  cohort, and coverage floor.} In panel A, cells give expected-region leads among
  correct verdicts under both availability rules and both cohorts. ``Strict''
  counts exams for which all three seeds agree on the expected region. The
  isolated cohort excludes concurrent positives among the three morphology
  labels 1dAVb, RBBB, and LBBB only. Panel B gives strict all-seed agreement
  under minimum beat-complete coverage floors; with no floor, it equals the
  strict beat-complete row of panel A. A floor changes the retained cohort, so
  the rows are not repeated measurements on one fixed set of exams.}
  \label{tab:si-cells}
  \fontsize{8}{10}\selectfont
  \textbf{A. Availability rule and cohort}\par\smallskip
  \begin{adjustbox}{max width=\linewidth,center}
  \begin{tabular}{llcccc}
    \toprule
    Rule & Cohort & 1dAVb & RBBB & LBBB & All \\
    \midrule
    beat-complete & all positive & 47/59 & 95/96 & 78/83 & 220/238 (92\%) \\
    per-region    & all positive & 45/59 & 96/96 & 82/83 & 223/238 (94\%) \\
    beat-complete & isolated     & 38/50 & 83/84 & 69/74 & 190/208 (91\%) \\
    per-region    & isolated     & 37/50 & 84/84 & 73/74 & 194/208 (93\%) \\
    \midrule
    \multicolumn{2}{l}{strict, beat-complete, all positive}
      & 16/28 & 32/34 & 26/30 & 74/92 \\
    \multicolumn{2}{l}{strict, per-region, all positive}
      & 14/28 & 34/34 & 28/30 & 76/92 \\
    \bottomrule
  \end{tabular}
  \end{adjustbox}

  \medskip
  \textbf{B. Coverage floors, strict beat-complete agreement}\par\smallskip
  \begin{adjustbox}{max width=\linewidth,center}
  \begin{tabular}{lccccc}
    \toprule
    Floor & Exam--class pairs kept & 1dAVb & RBBB & LBBB & All \\
    \midrule
    0.5  & 91 & 16/28 & 32/34 & 25/29 & 73/91 \\
    0.6  & 88 & 16/28 & 31/33 & 25/27 & 72/88 \\
    0.7  & 87 & 16/28 & 31/33 & 24/26 & 71/87 \\
    0.8  & 81 & 16/28 & 29/31 & 20/22 & 65/81 \\
    \bottomrule
  \end{tabular}
  \end{adjustbox}
\end{table}

\begin{table}
  \renewcommand{\arraystretch}{1.4}
  \centering
  \caption{\textbf{Beat-sequence interventions with 95\% exam-bootstrap intervals.}
  Cells give the fraction of label-positive verdicts that stay positive.
  ``Template, RR'' repeats the median-duration beat but restores each beat's
  original duration.}
  \label{tab:si-timing}
  \fontsize{8}{10}\selectfont
  \begin{adjustbox}{max width=\linewidth,center}
  \begin{tabular}{llccccc}
    \toprule
    Dataset & Label & Intact & Shuffled & Template & Template, RR & Equalized \\
    \midrule
    CODE-Test & 1dAVb & 0.70 [0.56, 0.85] & 0.65 [0.49, 0.80] & 0.68 [0.55, 0.81] & 0.68 [0.54, 0.82] & 0.70 [0.55, 0.85] \\
              & RBBB  & 0.94 [0.85, 1.00] & 0.95 [0.87, 1.00] & 0.89 [0.79, 0.97] & 0.88 [0.78, 0.97] & 0.95 [0.87, 1.00] \\
              & LBBB  & 0.92 [0.82, 1.00] & 0.90 [0.81, 0.98] & 0.94 [0.88, 1.00] & 0.94 [0.88, 1.00] & 0.92 [0.83, 1.00] \\
              & SB    & 1.00 [1.00, 1.00] & 0.97 [0.91, 1.00] & 0.76 [0.58, 0.93] & 0.73 [0.54, 0.90] & 0.92 [0.81, 1.00] \\
              & AF    & 1.00 [1.00, 1.00] & 1.00 [1.00, 1.00] & 0.00 [0.00, 0.00] & 0.00 [0.00, 0.00] & 1.00 [1.00, 1.00] \\
              & ST    & 0.91 [0.82, 0.98] & 0.91 [0.83, 0.98] & 0.92 [0.83, 0.99] & 0.92 [0.82, 0.99] & 0.91 [0.82, 0.98] \\
    \midrule
    PTB-XL    & 1dAVb & 0.61 [0.55, 0.67] & 0.60 [0.53, 0.66] & 0.55 [0.49, 0.62] & 0.55 [0.49, 0.61] & 0.59 [0.53, 0.65] \\
              & RBBB  & 0.95 [0.93, 0.98] & 0.96 [0.94, 0.98] & 0.87 [0.83, 0.91] & 0.87 [0.83, 0.91] & 0.95 [0.92, 0.97] \\
              & LBBB  & 0.94 [0.91, 0.97] & 0.94 [0.91, 0.97] & 0.92 [0.88, 0.95] & 0.92 [0.88, 0.95] & 0.94 [0.90, 0.97] \\
              & SB    & 0.67 [0.63, 0.70] & 0.65 [0.61, 0.69] & 0.56 [0.52, 0.59] & 0.54 [0.51, 0.58] & 0.62 [0.59, 0.66] \\
              & AF    & 0.94 [0.93, 0.95] & 0.94 [0.92, 0.95] & 0.01 [0.00, 0.01] & 0.01 [0.01, 0.01] & 0.93 [0.92, 0.94] \\
              & ST    & 0.94 [0.93, 0.96] & 0.93 [0.92, 0.95] & 0.94 [0.93, 0.95] & 0.94 [0.93, 0.95] & 0.94 [0.93, 0.96] \\
    \bottomrule
  \end{tabular}
  \end{adjustbox}
\end{table}

\begin{table}
  \renewcommand{\arraystretch}{1.4}
  \centering
  \caption{\textbf{Uniform and interval-preserving rate change on CODE-Test.} The
  crossing is the imposed heart rate at which half of the cohort's verdicts
  change call, with 95\% exam-bootstrap intervals. Interval-preserving rows use
  verdicts that remain feasible over the whole window. The lower block gives the
  fraction of conduction-positive verdicts that stay positive.}
  \label{tab:si-rate}
  \fontsize{8}{10}\selectfont
  \begin{adjustbox}{max width=\linewidth,center}
  \begin{tabular}{llcc}
    \toprule
    Cohort & Quantity & Uniform & Interval-preserving \\
    \midrule
    No target label & SB crossing (beats/min) & 47.8 [47.6, 48.0] & 45.4 [44.4, 45.8] \\
    No target label & ST crossing (beats/min) & 105.6 [104.9, 106.5] & infeasible \\
    ST positive & ST crossing (beats/min) & 99.4 [98.7, 100.5] & 101.9 [100.6, 102.7] \\
    SB positive & still SB at 60 beats/min & 0.00 & 0.00 \\
    ST positive & still ST at 90 beats/min & 0.00 & 0.00 \\
    \midrule
    1dAVb positive & 60 / 80 / 100 beats/min & 0.77 / 0.20 / 0.00 & 0.64 / 0.58 / infeasible \\
    RBBB positive  & 60 / 80 / 100 beats/min & 0.92 / 0.84 / 0.49 & 0.93 / 0.94 / 0.92 \\
    LBBB positive  & 60 / 80 / 100 beats/min & 0.97 / 0.83 / 0.47 & 0.81 / 0.89 / 1.00$^{*}$ \\
    \bottomrule
  \end{tabular}
  \end{adjustbox}
  \par\smallskip{\footnotesize $^{*}$At 100~beats/min, 24\% of RBBB and 7\% of LBBB verdicts are feasible.}
\end{table}

\begin{table}
  \renewcommand{\arraystretch}{1.4}
  \centering
  \caption{\textbf{Fraction of correct verdicts in which one lead alone keeps the call
  positive.} Top block CODE-Test, bottom block strict-SCP PTB-XL.}
  \label{tab:si-leads}
  \fontsize{8}{10}\selectfont
  \begin{adjustbox}{max width=\linewidth,center}
  \begin{tabular}{lcccccccccccc}
    \toprule
    Label & I & II & III & aVR & aVL & aVF & V1 & V2 & V3 & V4 & V5 & V6 \\
    \midrule
    1dAVb & 0.22 & 0.97 & 0.37 & 0.08 & 0.68 & 0.00 & 0.02 & 0.07 & 0.47 & 0.56 & 0.44 & 0.46 \\
    RBBB  & 0.95 & 0.97 & 0.48 & 0.54 & 0.81 & 0.27 & 0.00 & 0.00 & 0.03 & 0.41 & 0.80 & 0.95 \\
    LBBB  & 0.20 & 0.08 & 0.51 & 0.39 & 0.00 & 0.34 & 0.41 & 0.96 & 0.86 & 0.48 & 0.08 & 0.25 \\
    SB    & 0.62 & 0.83 & 0.46 & 0.69 & 0.56 & 0.17 & 0.06 & 0.02 & 0.58 & 0.50 & 0.60 & 0.50 \\
    AF    & 0.92 & 0.79 & 0.36 & 0.08 & 0.49 & 0.13 & 0.00 & 0.33 & 0.90 & 1.00 & 0.85 & 0.95 \\
    ST    & 0.79 & 0.76 & 0.47 & 0.36 & 0.54 & 0.11 & 0.00 & 0.22 & 0.61 & 0.80 & 0.61 & 0.76 \\
    \midrule
    1dAVb & 0.32 & 0.89 & 0.50 & 0.01 & 0.09 & 0.69 & 0.12 & 0.27 & 0.68 & 0.54 & 0.52 & 0.38 \\
    RBBB  & 0.96 & 0.89 & 0.47 & 0.33 & 0.96 & 0.51 & 0.00 & 0.00 & 0.01 & 0.42 & 0.85 & 0.96 \\
    LBBB  & 0.25 & 0.11 & 0.45 & 0.07 & 0.05 & 0.35 & 0.60 & 0.98 & 0.81 & 0.37 & 0.16 & 0.40 \\
    SB    & 0.68 & 0.68 & 0.36 & 0.37 & 0.35 & 0.39 & 0.01 & 0.02 & 0.65 & 0.56 & 0.61 & 0.42 \\
    AF    & 0.82 & 0.85 & 0.30 & 0.03 & 0.59 & 0.30 & 0.03 & 0.44 & 0.78 & 0.96 & 0.95 & 0.98 \\
    ST    & 0.80 & 0.71 & 0.38 & 0.01 & 0.52 & 0.62 & 0.00 & 0.23 & 0.64 & 0.77 & 0.66 & 0.71 \\
    \bottomrule
  \end{tabular}
  \end{adjustbox}
\end{table}

\begin{table}
  \renewcommand{\arraystretch}{1.4}
  \centering
  \caption{\textbf{Hyperparameter search space and selected configuration shared by
  runs with different random seeds.} The selected configuration has the highest
  macro F1 on the CODE100 validation split. Ranges are on a linear scale unless
  marked ``log''. ``Integer'' marks integer-valued ranges, and braces list
  categorical choices.}
  \label{tab:hpo}
  \fontsize{8}{10}\selectfont
  \begin{adjustbox}{max width=\linewidth,center}
  \begin{tabular}{lll}
    \toprule
    Hyperparameter & Search range & Selected \\
    \midrule
    Dropout              & $[0.0,\,0.15]$                          & 0.012 \\
    Head dropout         & $[0.0,\,0.15]$                          & 0.118 \\
    Weight decay         & $[10^{-5},\,10^{-3}]$, log              & \num{7.52e-4} \\
    Grad-clip norm       & $[0.5,\,2.0]$                           & 1.523 \\
    $n_{\text{neg}}$     & $[\num{150000},\,\num{500000}]$, log    & \num{220514} \\
    Learning rate        & $[10^{-4},\,5\times10^{-3}]$, log       & \num{1.36e-3} \\
    Batch size           & $\{64,\,128,\,256\}$                    & 128 \\
    Scheduler factor     & $[0.1,\,0.9]$                           & 0.589 \\
    Max epochs           & $[20,\,60]$, integer                    & 40 \\
    Early-stop patience  & $[5,\,20]$, integer                     & 15 \\
    \bottomrule
  \end{tabular}
  \end{adjustbox}
\end{table}
\FloatBarrier

\bibliographystyle{sn-nature}
\bibliography{refs}